\documentclass[letterpaper, 10 pt, journal, twoside]{IEEEtran}
\usepackage{amsmath,amsfonts}
\usepackage{array}
\usepackage[caption=false,font=normalsize,labelfont=sf,textfont=sf]{subfig}
\usepackage{textcomp}
\usepackage{stfloats}
\usepackage{url}
\usepackage{verbatim}
\usepackage{graphicx}
\usepackage{multirow}
\usepackage{amsthm}

\usepackage{bm}
\usepackage{color}
\usepackage{nomencl}
\usepackage[linesnumbered,ruled,vlined]{algorithm2e}
\usepackage[hidelinks]{hyperref}
\newcommand{\citeblue}[1]{{\textcolor{blue}{\cite{#1}}}}

\usepackage{amssymb}
\usepackage[noend]{algpseudocode}

\usepackage{float}
\usepackage{tabularx}
\usepackage{booktabs}
\usepackage{tikz}

\makenomenclature
\def\BibTeX{{\rm B\kern-.05em{\sc i\kern-.025em b}\kern-.08em
    T\kern-.1667em\lower.7ex\hbox{E}\kern-.125emX}}
\usepackage{balance}

\newcommand{\wrapExpr}[1]{${#1}$\xspace}

\newcommand{\numValid}{N_\text{valid}}
\newcommand{\numInvalid}{{N_\text{invalid}}}
\newcommand{\compF}{\vec{F}_{\hat{\mathcal{D}}}}
\newcommand{\repuF}{\vec{F}_{\text{repulsive}}}
\newcommand{\attrF}{\vec{F}_{\text{attractive}}}
\newcommand{\batchSize}{\mathcal{B}}
\newcommand{\decayFac}{\Theta_\textit{d}}
\newcommand{\smoothingFac}{\mathcal{\sigma}_\textit{s}}
\newcommand{\tuningFac}{\mathcal{\tau}_\textit{t}}
\newcommand{\ratioLeb}{\mathcal{G}_i}
\newcommand{\leb}{\mathcal{\zeta}}
\newcommand{\charge}{q}

\newcommand{\numValidT}{\wrapExpr{\numValid}}
\newcommand{\numInvalidT}{\wrapExpr{\numInvalid}}
\newcommand{\compFT}{\wrapExpr{\compF}}
\newcommand{\repuFT}{\wrapExpr{\repuF}}
\newcommand{\attrFT}{\wrapExpr{\attrF}}
\newcommand{\batchSizeT}{\wrapExpr{\batchSize}}
\newcommand{\decayFacT}{\wrapExpr{\decayFac}}
\newcommand{\smoothingFacT}{\wrapExpr{\smoothingFac}}
\newcommand{\tuningFacT}{\wrapExpr{\tuningFac}}
\newcommand{\ratioLebT}{\wrapExpr{\ratioLeb}}
\newcommand{\lebT}{\wrapExpr{\leb}}
\newcommand{\chargeT}{\wrapExpr{\charge}}

\begin{document}
\title{APT*: Asymptotically Optimal Motion Planning via Adaptively Prolated Elliptical R-Nearest Neighbors}
\author{Liding Zhang$^{1}$, Sicheng Wang$^{1}$, Kuanqi Cai$^{1}$, Zhenshan Bing$^{2,1}$, Fan Wu$^{1}$, Chaoqun Wang$^{3}$, \\Sami Haddadin$^{1}$, ~\IEEEmembership{Fellow,~IEEE}, Alois Knoll$^{1}$, ~\IEEEmembership{Fellow,~IEEE} % <-this % stops a space

\thanks{Manuscript received: May, 29, 2024; Accepted July, 31, 2025. This paper was recommended for publication by Editor Júlia Borràs Sol upon evaluation of the Associate Editor and Reviewers’ comments.
\textit{(Corresponding author: Zhenshan Bing.)}
} %Use only for final RAL version
\thanks{$^{1}$L. Zhang, S. Wang, K. Cai, Z. Bing, F. Wu, S. Haddadin and A. Knoll are with the School of Computation, Information and Technology (CIT), Technical University of Munich, 80333 Munich, Germany. {\tt\small liding.zhang@tum.de}}

\thanks{$^{2}$Zhenshan Bing is also with the State Key Laboratory for Novel Software Technology and the School of Science and Technology, Nanjing University (Suzhou Campus), China.}
\thanks{$^{3}$C. Wang is with the School of Control Science and Engineering, Shandong University, Shandong, China.}
\thanks{Digital Object Identifier (DOI): see top of this page.}
}
\markboth{IEEE Robotics and Automation Letters. Preprint Version. Accepted July, 2025}
{Zhang \MakeLowercase{\textit{et al.}}: APT*: Asymptotically Optimal Motion Planning via Adaptively Prolated Elliptical R-Nearest Neighbors} 

\maketitle

\begin{abstract}
Optimal path planning aims to determine a sequence of states from a start to a goal while accounting for planning objectives. Popular methods often integrate fixed batch sizes and neglect information on obstacles, which is not problem-specific. \textcolor{black}{This study introduces Adaptively Prolated Trees (APT*), a novel sampling-based motion planner that extends based on Force Direction Informed Trees (FDIT*), integrating adaptive batch-sizing and elliptical $r$-nearest neighbor modules to dynamically modulate the path searching process based on environmental feedback.} APT* adjusts batch sizes based on the hypervolume of the informed sets and considers vertices as electric charges that obey Coulomb’s law to define virtual forces via neighbor samples, thereby refining the prolate nearest neighbor selection.
These modules employ non-linear prolate methods to adaptively adjust the electric charges of vertices for force definition, thereby improving the convergence rate with lower solution costs. Comparative analyses show that APT* outperforms existing single-query sampling-based planners in dimensions from $\mathbb{R}^4$ to $\mathbb{R}^{16}$, and it was further validated through a real-world robot manipulation task.
A video showcasing our experimental results is available at: \href{https://youtu.be/gCcUr8LiEw4}{\textcolor{black}{https://youtu.be/gCcUr8LiEw4}}.
\end{abstract}


\section{Introduction}
\IEEEPARstart{R}{obot} motion planning focuses on computing collision-free paths between start and goal configurations while avoiding obstacles~\citeblue{gammell2021asymptotically}. In high-dimensional spaces, sampling-based algorithms~\citeblue{Orthey2023AnnualReview} are widely regarded as indispensable for solving \textit{the curse of dimensionality} problem~\citeblue{zhang2024review}. These algorithms are underpinned by the theory of \textit{random geometric graphs} (RGG)~\citeblue{penrose2003random}, a probabilistic framework modeling randomly distributed networks within the \textit{configuration space} (\textit{$\mathcal{C}$-space}).
%
% RGGs place samples randomly in \textit{$\mathcal{C}$-space}, connecting pairs of samples within a predefined radius. 
The RGG's structure is characterized by random sample distribution in $n$-dimensional \textit{$\mathcal{C}$-space} within a predefined radius and proximity-based edge formation, typically using Euclidean metrics. This ensures asymptotic optimality in path planning with increased sample density~\citeblue{zhang2025TASE}.
\subsection{Related Work}
Search-based algorithms like Dijkstra's~\citeblue{dijkstra1959note} compute shortest paths by exhaustively exploring all routes, while A*~\citeblue{hart1968formal} improves efficiency with heuristic guidance. Sampling-based planners, such as Rapidly-exploring Random Trees (RRT)~\citeblue{lavalle1998rapidly} and Probabilistic Roadmaps (PRM)~\citeblue{kavraki1996probabilistic}, excel in handling non-convex and high-dimensional spaces.
RRT grows a tree incrementally from the start toward the goal, where RRT-Connect~\citeblue{connect2000} enhances efficiency by growing trees from both ends. RRT*~\citeblue{karaman2011sampling} improves this by using \textit{$k$-nearest}~\citeblue{fmt2015} or \textit{$r$-nearest}~\citeblue{Zhang2024Elliptical} neighbor searches and rewiring, offering anytime solutions with guaranteed optimality. Informed RRT*~\citeblue{gammell2014informed} accelerates convergence rate and path optimization by limiting the search space via elliptical \textit{informed sampling} subsets~\citeblue{gammell2018informed}.

Batch Informed Trees (BIT*)~\citeblue{gammell2020batch} integrates Informed RRT* and Fast Marching Trees (FMT*)~\citeblue{fmt2015}, advancing sampling techniques to group states into a compact \textit{batch}-based implicit RGG. 
% Its advanced variant, ABIT*~\citeblue{strub2020advanced}, introduces inflation and truncation factors to balance exploration and exploitation in dense RGG approximations.
%
Adaptively Informed Trees (AIT*)~\citeblue{strub2022adaptively} and Effort Informed Trees (EIT*)~\citeblue{strub2022adaptively} employ asymmetrical \textit{forward-reverse} search with sparse collision checks in reverse search, enhancing exploration efficiency. Flexible Informed Trees (FIT*)~\citeblue{Zhang2024adaptive} updates the batch size of each improved approximation for faster initial convergence. \textcolor{black}{Force Direction Informed Trees (FDIT*)~\citeblue{Zhang2024Elliptical} builds upon EIT* and utilizes Coulomb's law with fixed vertex charge to calculate forces that define elliptical neighbor regions.}
However, in Coulomb's law, the magnitude of the force is determined not only by the distance between particles but also by their charges. When applied to path planning, it is crucial to optimize the vertices charge regarding planning phases~\citeblue{zhang25ral}. Prolating neighbors too early may lead to local optimality, while prolate neighbors too late may result in the loss of problem-specific information.

\subsection{Proposed Algorithm and Original Contribution}
\begin{figure*}[t!] 
    \centering
    \begin{tikzpicture}

    \node[inner sep=0pt] (russell) at (0,0)
    {\includegraphics[width=0.99\textwidth]{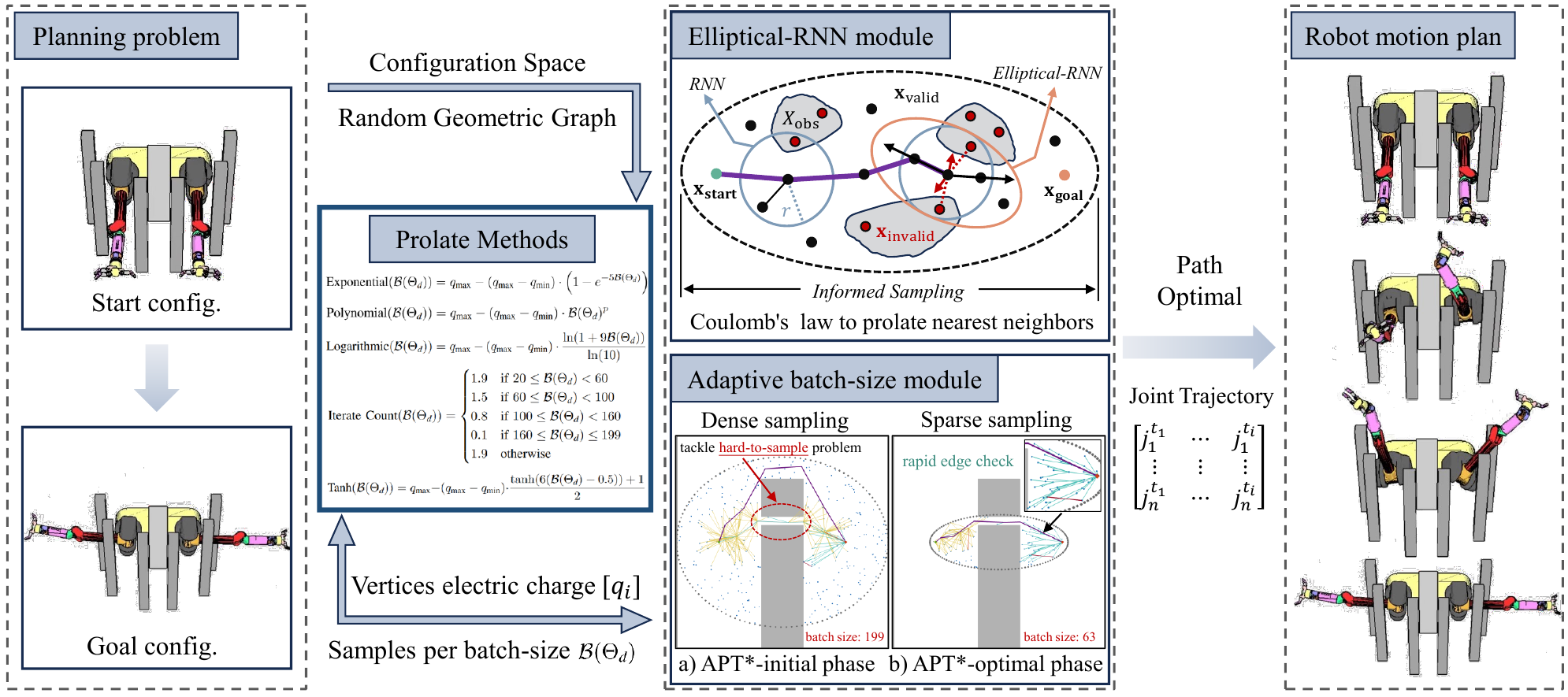}};

    \end{tikzpicture}
    \vspace{-0.7em} 
\caption{System diagram of the proposed adaptively prolated method. After defining the start and goal configurations, APT* dynamically adjusts the prolated search region. The framework comprises two primary modules: the Elliptical-RNN module (Fig.~\ref{fig:neighbor_compare}), which leverages Coulomb's law to compute the forces of valid and invalid nearest neighbor vertices, and the adaptive batch-size module, which optimizes the number of samples per batch based on the Lebesgue measure of the informed sets. These mechanisms ensure that the batch size is adaptively tuned according to the current phase of the search process.}
\label{fig: illstra}
\vspace{-1.7em} 
\end{figure*}

This paper introduces Adaptively Prolated Trees (APT*), an extension of the sampling-based planner FDIT*, which employs non-linear prolation strategies to dynamically adjust nearest neighbor regions using the adaptive batch-size module. In this module, dense sampling corresponds to the initial phase with a smaller charge, leading to more circular neighbors, while sparse sampling indicates that the path optimization has progressed, allowing a larger charge to increase the Coulomb force's magnitudes, thereby making the neighbors more elliptical. This adjustment impacts the Coulomb force calculation, which is applied to the elliptical $r$-nearest neighbors (elliptical-RNN) module (Fig.~\ref{fig: illstra}), improving search and rewiring efficiency. The approach utilizes the Taylor series expansion of \textit{Tanh} function to modulate sample charges, directing the search toward valuable regions and enhancing overall pathfinding performance.
% APT* extends the fixed-force elliptical-RNN search by incorporating adaptive prolating Coulomb’s law-based force direction analysis, which guides the search and avoids costly or infeasible areas. 
% This enhances search efficiency and reduces path costs, particularly in high-dimensional, multi-narrow passage scenarios. 
% APT* was tested on the dual-manipulator simulation and the real-world robot tasks, demonstrating its effectiveness in high-dimensional, multi-narrow passage scenarios.
%
%
\textcolor{black}{
This work contributes the following extensions to FDIT* and EIT*:
\begin{enumerate}
    \item A non-linear prolate method based on an adaptive batch size module to affect vertices' electric charges, optimizing the Coulomb forces during the planning phase.
    \item A novel sampling-based path planner APT* that integrates adaptively prolated elliptical-RNN for search and rewiring to quickly obtain high-quality solutions.
    \item Demonstration of APT*'s effectiveness across various dimensional environments and optimization objectives.
\end{enumerate}}
\section{Problem Formulation and Notation}
% In this section, we first formulate the problem of optimal path planning and then introduce the notations in APT*.
\subsection{Problem Definition}
\textit{Optimal Planning:} Given a planning problem with state space $X \subseteq \mathbb{R}^n$, where $X_{\text{obs}}$ represents the states in collision with obstacles, and $X_{\text{free}} = cl(X \setminus X_{\text{obs}})$ denotes the set of permissible states, with $cl(\cdot)$ indicating the \textit{closure} of a set. The initial state is $\mathbf{x}_{\text{start}} \in X_{\text{free}}$, and the goal states are in $X_{\text{goal}} \subset X_{\text{free}}$. A continuous map $\sigma: [0, 1] \mapsto X$ represents a collision-free path, and $\Sigma$ is the set of all nontrivial paths~\citeblue{karaman2011sampling}.

The optimal solution, $\sigma^*$, is the path that minimizes a chosen cost function $c: \Sigma \mapsto \mathbb{R}_{\geq 0}$. This path connects the initial state $\mathbf{x}_{\text{start}}$ to a goal state $\mathbf{x}_{\text{goal}} \in X_{\text{goal}}$ through the free space:
\begin{equation}
\begin{split}
    \sigma^* &= \arg \min_{\sigma \in \Sigma} \left\{ c(\sigma) \middle| \sigma(0) = \mathbf{x}_{\text{start}}, \sigma(1) \in \mathbf{x}_{\text{goal}}, \right. \\
    &\qquad\qquad \left. \forall t \in [0, 1], \sigma(t) \in X_{\text{free}} \right\}.
\end{split}
\end{equation}
where $\mathbb{R}_{\geq 0}$ denotes the set of non-negative real numbers. The cost of the optimal path is denoted as $c^*$.
When considering a discrete set of states $X_{\text{samples}} \subset X$, represented as a graph with edges determined by a transition function, its properties can be modeled probabilistically using implicit dense RGGs, where $X_{\text{samples}} = \{\mathbf{x} \sim \mathcal{U}(X)\}$, as discussed in~\citeblue{penrose2003random}.
\subsection{Notation}~\label{subsec: notation}
The state space of the planning problem is denoted by \( X \subseteq \mathbb{R}^n \), where \( n \in \mathbb{N} \). The start vertex is represented by \( \mathbf{x}_{\text{start}} \in X \), and the goals are denoted by \( X_{\text{goal}} \subset X \). The coordinates for valid and invalid vertices are given by \( \mathbf{x}_{\text{valid}} \) and \( \mathbf{x}_{\text{invalid}} \). 
% The forward and reverse search trees are represented by \( \mathcal{F} \) and \( \mathcal{R} \), respectively. The vertices in these trees, denoted by \( V_\mathcal{F} \) and \( V_\mathcal{R} \), correspond to valid states. 
An admissible estimate for the cost of a path is \( \hat{f}: X \rightarrow [0, \infty) \), which characterizes the informed set \( X_{\hat{f}} \).

\textit{APT*-specific Notation:}
We assume each vertex has properties of electric charge, denoted as \chargeT. These vertices generate either an attractive force, denoted as \attrFT, or a repulsive force, denoted as \repuFT. The Euclidean distance between two vertices is given by \(r_i\), and the unit direction vector between them is denoted by \(\mathbf{\hat{r}_i}\). The total virtual Coulomb force acting on the current processing vertex from neighbors, represented as \compFT, is calculated using Coulomb's law and incorporates the elliptical search neighbors, denoted as \(V_\textit{ellipseNeighbors}\). Within this set of neighbors, the ratio of invalid vertices is represented by \(\Phi\), while the number of valid and invalid vertices are denoted by \numValidT and \numInvalidT, respectively.

\textcolor{black}{The batch sizes are denoted as $\batchSize(\decayFac)$, which is tuned by a decay factor \decayFacT, with specified minimum \( m_\text{min} \) and maximum \( m_\text{max} \) sample numbers per batch.} The Lebesgue measure (i.e., $n$-dimensional hypervolume) of a prolated hyperellipsoid is represented by $\leb (c_i, n)$.
The non-negative scalar \ratioLebT represents the informed ratio of the Lebesgue measure of hyperellipsoids. \tuningFacT regulates the rate at which the ratio decreases. A sigmoid-based smoothing parameter \smoothingFacT represents a smoothed value after the decay of the initial and optimal states.

\section{Adaptively Prolated Trees (APT*)}
\begin{algorithm}[t!]
\caption{{APT* - Elliptical $r$-nearest neighbors}}
\label{alg:elliptcal_search}
\small
\DontPrintSemicolon
\SetKwInOut{Input}{Input}
\SetKwInOut{Output}{Output}
\SetKwFunction{IsValid}{isValid}
\SetKwFunction{EllipseNeighbors}{getEllipseNeighbors}
\SetKwFunction{CalculateRatio}{calRatio}
\SetKwFunction{FDistance}{distance}
\SetKwFunction{InEllipse}{inEllipse}

\SetKwFunction{GetCharge}{getVertexCharge}
\SetKwFunction{GetRadius}{getRNNRadius}
\SetKwFunction{ProlateDistance}{calProlateDistance}
\Input{$\mathbf{x}$ - The current state, $\batchSize_\text{adapt}$ - The adapted batch size, \( n \) - The dimensionality}
\Output{Best set of nearest neighbors $V_\textit{{ellipseNeighbors}}$ within an ellipsoidal region around $\mathbf{x}$}

\BlankLine
%\( \textit{totalVertices} \gets 0 \) , \( %\textit{invalidVertices} \gets 0 \)\;
$V_\textit{ellipseNeighbors} \gets X_\text{samples}$\Comment{initialize all neighbors}\\
$\compF \gets \Vec{1},\Phi \leftarrow 1, r \leftarrow \GetRadius(\batchSize_\text{adapt})$\Comment{Eq.~\ref{eqn:radius}}\\

%$V_\textit{ellipseNeighbors} \gets \EllipseNeighbors(\mathbf{x}, r, \compF)$\\
% $\Phi \gets \CalculateRatio(V_\textit{ellipseNeighbors})$\\ \Comment{initialize charge ratio, skip loop when 90\% NN is valid}
\While(\Comment{neighbor invalid samples less than 10\%}){$\Phi \geq 0.1$}{

\ForEach(\Comment{update \compFT}){$\mathbf{x}_\textit{i} \in V_\textit{ellipseNeighbors}$}{
        % $\attrF \leftarrow \Vec{0}$, 
        % $\repuF \leftarrow \Vec{0}$\\
     $r_\textit{i} \gets \FDistance(\mathbf{x}, \mathbf{x}_\textit{i})$ \\
     %\Comment{compute the states distance}\\
      $\mathbf{\hat{r}}_\textit{i} \gets(\mathbf{x}_\textit{i} - \mathbf{x})/{r_\textit{i}}$  \Comment{compute the direction of $\mathbf{x}_\textit{i}$}\\
      \textcolor{black}{$\charge_\text{i} \leftarrow \GetCharge(\batchSize_\text{adapt})$}\Comment{Alg.~\ref{alg: nonlinearProlateMethods}} \\
      \eIf{\IsValid{$\mathbf{x}_\textit{i}$}}{
       
        \emph{\color{black}$\attrF \stackrel{}{\leftarrow}  \frac{\charge_\text{i}^{2}}{r_\textit{i}^{n-1}} \mathbf{\hat{r}}_\textit{i}$ \label{line:forceAttra}}\Comment{attractive positive force}\\
      }{
 \emph{\color{black}$\repuF \stackrel{}{\leftarrow} - \frac{\charge_\text{i}^{2}}{r_\textit{i}^{n-1}} \mathbf{\hat{r}}_\textit{i}$} \Comment{repulsive negative force}\\
      }
    $\compF \leftarrow \compF + \attrF +\repuF $ \Comment{Eq.~\ref{fuc:totalForcevector}}
    }
    % \emph{\color{black}$V_{\textit{ellipseNeighbors}} \gets \EllipseNeighbors(\mathbf{x}, k, \compF)$}\\ \Comment{recalculate NN based on the updated force}
  
    $N_\text{total} \leftarrow 0, \numInvalid \leftarrow 0$\\
    \ForEach(){\(\mathbf{x}_\textit{i} \in V_\textit{ellipseNeighbors}\)}{

    % \textcolor{black}{$d_p \leftarrow \ProlateDistance(\mathbf{x},\mathbf{x}_\textit{i},\compF)$}\\
    \eIf(\Comment{Eq.~\ref{fuc:elliptical-RNN}}){$\InEllipse(\mathbf{x}, \mathbf{x}_\textit{i}, \compF, n) $}{
        \( N_\text{total} \gets N_\text{total} + 1 \)\;
        \If{\textbf{not} $\IsValid(\mathbf{x}_\textit{i})$}{
            \( \numInvalid \gets \numInvalid + 1 \)\;
        }
    
    }{

        $V_\textit{ellipseNeighbors}{\leftarrow}V_\textit{ellipseNeighbors}\setminus{\mathbf{x}_\textit{i}}$
    }
    }
            \( \Phi \gets {\numInvalid}/{N_\text{total}} \) \Comment{non-negative charge ratio $\in$ $(0,1]$}
    % \eIf{\(\textit{totalVertices} > 0\)}{
       
    % }{
    %     \( \Phi \gets 0 \)\Comment{no samples in nearest neighbor}
    % }
}

\ForEach{$\mathbf{x}_\textit{i} \in V_\textit{{ellipseNeighbors}}$}{
    \If{ not \IsValid{$\mathbf{x}_\textit{i}$}}{$
        V_\textit{ellipseNeighbors}{\leftarrow}V_\textit{ellipseNeighbors}\setminus{\mathbf{x}_\textit{i}}$
    }
}
\Return $V_\textit{{ellipseNeighbors}}$
\end{algorithm}
In this section, we first introduce the concept of the Elliptical-RNN module. Next, we illustrate the adaptive batch-size module. Finally, we propose and evaluate nearest neighbor prolated methods based on non-linear functions.
\subsection{Elliptical R-Nearest Neighbors (RNN) Module}\label{subsec:elliptical RNN}
%\begin{proof}
%\textit{Extension of Coulomb's law to high %dimensions:}
%
Popular path planners use RNN to traverse the neighbors around the current state, and the radius of nearest neighbors \( r(\batchSize) \), crucial for defining the graph's sparsity, defined as:
    \begin{equation}
    \label{eqn:radius}
        r(\batchSize) := 2\eta \left(\left(1 + \frac{1}{n}\right){\left(\frac{\lambda(X_{\hat{f}})}{\lambda\left(U_{B, n}\right)}\right) \left( \frac{\log(\batchSize)}{\batchSize}\right)}\right)^{\frac{1}{n}},
    \end{equation}
where $\eta$ is a normalization constant, $n$ represents the dimensionality of the space, $\lambda(\cdot)$ denotes the Lebesgue measure, $\batchSize$ is the current batch size, and $U_{B, n}$ is a unit ball in an $n$-dimensional space~\citeblue{strub2022adaptively}.
This formula ensures that $r(\batchSize)$ adapts to the complexity of the space, enhancing the graph's connectivity without increasing computational effort~\citeblue{solovey2020critical}.
\begin{figure}[t!]
    \centering
    \begin{tikzpicture}
    \node[anchor=center] at (0,0) 
    {\includegraphics[width=0.48\textwidth]{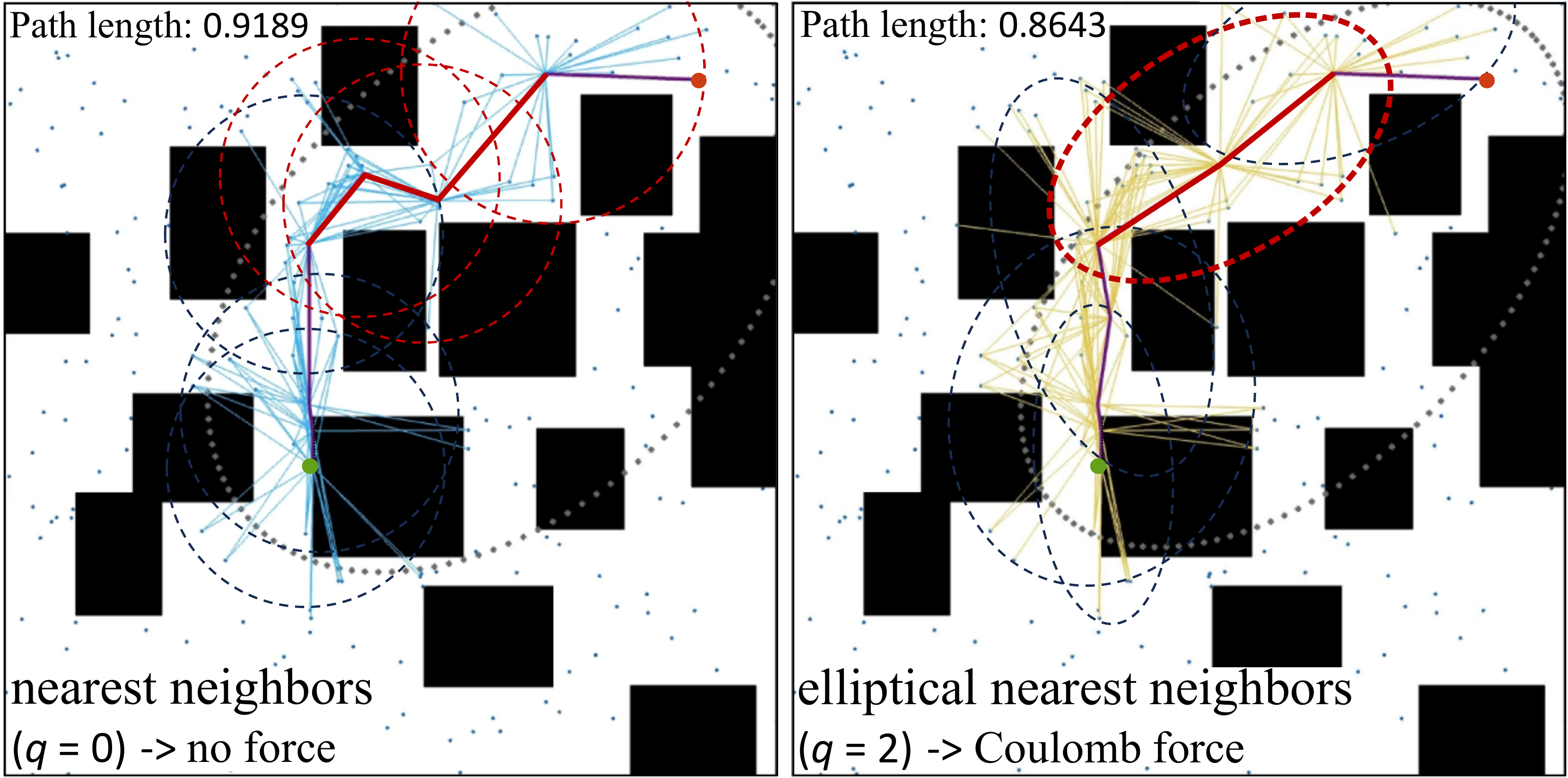}};
    \end{tikzpicture}
    \vspace{-2.0em}
    \caption{\textcolor{black}{Illustrations of the 2D representation of elliptical nearest neighbor search. The vertex charge $q$ in Coulomb’s law affects the force exerted on elliptical nearest neighbors during the exploration and rewiring phases.}}
    \label{fig:neighbor_compare}
    \vspace{-1.7em}
\end{figure}

We proposed an adaptively prolated method based on the traditional RNN via Coulomb's law. Consider a \(n\)-dimensional Euclidean space \( \mathbb{R}^n \) (Fig.~\ref{fig:neighbor_compare}). Let two vertexes in \textit{$\mathcal{C}$-space} with charge $\charge$, separated by a distance \( r_\textit{i} \). The unit vector from current vertex $\mathbf{x}$ to vertex $\mathbf{x}_\textit{i}$ is denoted by \( \mathbf{\hat{r}}_\textit{i} \). The \(n\)-dimensional Coulomb force is calculated as:
\begin{equation}
\compF := \sum_{i=1}^{N_\text{total}}\pm\mathbf{k}_e \frac{ \charge^{2}}{r_\text{i}^{n-1}} \mathbf{\hat{r}_\text{i}},
\end{equation}
where \compFT represents the force vector exerted by the charges on each other, and $N_\text{total}$ is the number of neighbors around the current state. 
% The proportionality constant \( \mathbf{k}_e \) is analogous to Coulomb's constant in three dimensions, determining the strength of the force. \( n \) represents the dimensionality of the \textit{$\mathcal{C}$-space} where these interactions occur.
%\end{proof}
%
For a given vertex, we have 
\numValidT valid vertices and 
\numInvalidT invalid vertices in its neighborhood. The vertices in free space exert an attractive force on the neighborhood of the vertex, represented as \attrFT. Conversely, the surrounding vertices located within obstacles exert a repulsive force on the neighborhood of the vertex, represented as \repuFT. These two force vectors can be obtained through the formula below:
\begin{equation}
\attrF(\mathbf{x}) = \mathbf{k}_e \sum_{i=1}^{\numValid} \frac{\charge^2}{\|\mathbf{x} - \mathbf{x}_{\text{valid}, i}\|^{n}} (\mathbf{x}_{\text{valid}, i} - \mathbf{x}),
\end{equation}
\begin{equation}
\repuF(\mathbf{x}) = -\mathbf{k}_e \sum_{j=1}^{\numInvalid} \frac{\charge^2}{\|\mathbf{x} - \mathbf{x}_{\text{invalid}, j}\|^{n}} (\mathbf{x}_{\text{invalid}, j} - \mathbf{x}).
\end{equation}
where \( \mathbf{k}_e \) is a proportionality constant to Coulomb's constant. $\charge$ is the charge of the current state, which is dynamically adjusted via the adaptive batch-size module. \( \mathbf{x} \) is the position vector of the current state, \( \mathbf{x}_{\text{valid}, i} \) and \( \mathbf{x}_{\text{invalid}, j} \) are the position vectors of the \( i \)-th valid vertex and the \( j \)-th invalid vertex, respectively. \( n \) is the dimensionality of the \textit{$\mathcal{C}$-space}.

\textcolor{black}{
For each vertex \( \mathbf{x} \) in the \textit{$\mathcal{C}$-space}, all other sampled vertices exert either attractive or repulsive Coulomb-like forces on it. The total force acting on the current vertex is computed as the vector sum of all these interactions:
\begin{equation}
\label{fuc:totalForcevector}
\compF(\mathbf{x}) = \attrF(\mathbf{x}) + \repuF(\mathbf{x}),
\end{equation}
under the influence of the resultant Coulomb force \(\compF(\mathbf{x})\), the conventional isotropic RNN ball centered at \(\mathbf{x}\) is elongated in the direction of \(\compF(\mathbf{x})\) and compressed in the orthogonal subspace. The resulting anisotropic neighborhood forms an ellipsoid whose major axis aligns with \(\compF(\mathbf{x})\); we refer to this deformed search region as the \textit{elliptical-RNN}. To obtain the major-axis direction, the Coulomb force vector is normalized as \( \mathbf{u}_1 = \frac{\compF(\mathbf{x})}{\|\compF(\mathbf{x})\|} \).
The Gram–Schmidt orthogonalization (i.e., QR decomposition)~\citeblue{gammell2018informed} is then applied to construct \(n - 1\) additional mutually orthogonal unit vectors \(\mathbf{u}_{2}, \ldots, \mathbf{u}_{n}\), resulting in the orthogonal matrix:
\begin{equation}
\label{fuc:orthogonalQ}
\mathbf{Q} = [\mathbf{u}_1\ \mathbf{u}_2\ \ldots\ \mathbf{u}_n] \in \mathbb{R}^{n \times n},
\end{equation}
therefore, the semi-axis is prolonged and aligned with \(\compF(\mathbf{x})\) to \(d_{1} = r\left(1 + k\|\compF\|\right)\), where \(k\) is a scaling factor, while keeping the remaining semi-axes at \(d_{2..n} = r\). These lengths are inserted into the diagonal scaling matrix:
\begin{equation}
\label{fuc:scalingD}
\mathbf{D}^{-2} = \mathrm{diag}(d_{1}^{-2}, r^{-2}, \ldots, r^{-2}),
\end{equation}
the elliptical-RNN region \(X_{\text{eRNN}}(\mathbf{x})\) centered at $\mathbf{x}$ is defined as the set of points $\mathbf{x}_i \in \mathbb{R}^n$ satisfying:
\begin{equation}
\label{fuc:elliptical-RNN}
X_{\text{eRNN}}(\mathbf{x}) := \left\{\mathbf{x}_\textit{i} \in \mathbb{R}^n \ \big| (\mathbf{x}_i-\mathbf{x})^\top\mathbf{Q}\mathbf{D}^{-2}\mathbf{Q}^\top(\mathbf{x}_i-\mathbf{x})<1\right\},
\end{equation}
where the inequality defines the interior of a hyperellipsoid centered at $\mathbf{x}$, aligned along the direction of the force.
}

%
% In contemplating the combined influence of numerous sample vertices on the present condition, these expressions prove versatile  within high-dimensional state spaces.
%
\subsection{Adaptive Batch-size Module}\label{subsec:adaptive-batch}
In this module, we establish the relationship between batch size and the hypervolume of informed sets. Intuitively, larger batch sizes facilitate faster discovery of an initial feasible path, while smaller batch sizes enhance optimization by minimizing edge checks and computational costs (Alg.~\ref{alg:adaptiveBatch}).
\textcolor{black}{The batch size \batchSizeT is dynamically adjusted each time by the following formula:
\begin{equation}
\batchSize(\decayFac) := \lfloor m_\text{min} + \decayFac \times (m_\text{max}-m_\text{min})\rfloor ,
\end{equation}
where $\lfloor \cdot \rfloor$ is the floor function, the decay factor $\decayFac \in (0,1)$ is logarithmically smoothed to prevent abrupt transitions. When $\decayFac$ approaches 1, $\batchSize$ nears the $m_\text{max}$, facilitating a faster initial solution. Conversely, when $\decayFac$ approaches 0, $\batchSize$ is closer to the $m_\text{min}$, promoting more efficient path optimization:
\begin{equation}\label{fuc:decay}
\decayFac = \frac{\ln(\tuningFac \times \smoothingFac +1)}{\ln{(\tuningFac +1)}}.
\end{equation}
where the tunning parameter $\tuningFac = (m_\text{max}+m_\text{min})/n_\text{dim}$ is related to the problem domain's dimensionality.} The smoothing parameter \smoothingFacT via the sigmoid function (without overflow form) is defined as:
\begin{equation}\label{fuc:smooth}
\smoothingFac = 
\begin{cases} 
{e^{10 (\ratioLeb - 0.5)}}/{(1 + e^{10  (\ratioLeb - 0.5)})}, & \text{if } \ratioLeb  < 0.5 \\
{1}/{(1 + e^{-10  (\ratioLeb - 0.5)})}, & \text{otherwise}
\end{cases},
\end{equation}
where $e$ is Euler’s number $e = \sum_{i=0}^\infty \frac{1}{n!}$. This ensures gradual transitions in batch sizes, correlating with informed ratio changes. The informed ratio \ratioLebT, described as:
\begin{equation}
\label{fuc:ratio}
    {\ratioLeb = \frac{\leb (c_\textit{current}, n)}{\leb (c_\textit{initial}, n)},}
\end{equation}
where \lebT ($\cdot, \cdot$) is the hypervolume contraction as solution cost improves to adjust the batch-size. This ratio \ratioLebT bounded within (0,1], it ensures batch sizes adapt to the problem’s optimization phase. The Lebesgue measure (i.e., hypervolume) of $n$-dimensional hyper ellipsoid $\leb (c_i, n)$ is defined as:
\begin{equation}
\leb (c_i , n) = \frac{\pi^{\frac{n}{2}}c_i\left(c_i^2-c_{\min}^2\right)^{\frac{n-1}2}}{2^n\cdot\Gamma\left(\frac{n}{2} + 1\right)}
.\end{equation}
where \(\Gamma(\cdot)\) is the gamma function, an extension of factorials to real numbers, $c_\text{min}$ denotes the Euclidean distance between $\mathbf{x}_{\text{start}}$ and $\mathbf{x}_{\text{goal}}$, $c_i$ is the current cost of solution.

\begin{algorithm}[t!]
\caption{\text{APT* - \textcolor{black}{Adaptive batch-size}}}
\label{alg:adaptiveBatch}
\DontPrintSemicolon
% \scriptsize
\small
\SetKwInOut{Input}{Input}
\SetKwInOut{Output}{Output}
\SetKwIF{If}{ElseIf}{Else}{if}{}{else if}{else}{end if}%
\SetKwFunction{calTuningParam}{calTuningParam}
\SetKwFunction{calDecayFac}{calDecayFac}
\SetKwFunction{sigmoidSmooth}{sigmoidSmooth}
\SetKwFunction{informedRatio}{informedRatio}
\SetKwFunction{collisionFree}{collisionFree}
\SetKwFunction{lebMeasure}{lebMeasure}
\SetKwFunction{isCostBetter}{isCostBetter}
\Input{last 
cost $c_\text{last}$, current cost $c_\text{current}$, dimensionality $n$}
\Output{$\batchSize_\text{adapt}$ - The adapted batch-size in informed set}
\emph{$m_\text{min}\leftarrow 1, m_\text{max} \leftarrow 2m_\text{current} - m_\text{min}$, $\batchSize_{\text{adapt}} \leftarrow0$}\\
% \emph{$\batchSize_{\text{adapt}} \leftarrow0$}\\
\emph{$\tuningFac \leftarrow \calTuningParam(m_\text{max},m_\text{min})$}\\
% \If {$c_\text{last} = \infty$}
% {
% \Return $\batchSize_\text{adapt}$\Comment{ensure safety when no initial solution}
% }
\If{$c_\text{current} \neq c_\text{last}\;\mathbf{or}\;c_\text{last} := \infty$}
{
    \emph{$c_\text{last} \leftarrow$ \isCostBetter($c_\text{current}$)}\\
    \If{pragma once}{\emph{$\leb_\text{initial} \leftarrow \lebMeasure (c_\text{last},n)$}\Comment{initial measure}\\}

    \emph{$\leb_\text{current}\leftarrow \lebMeasure(c_\text{current},n)$} 
    \Comment{updated measure}\\
    \emph{$\ratioLeb \leftarrow \informedRatio(\leb_\text{current},\leb_\text{initial})$}\Comment{Eq.~\ref{fuc:ratio}}\\
    \emph{$\smoothingFac \leftarrow \sigmoidSmooth(\ratioLeb)$}
    \Comment{Eq.~\ref{fuc:smooth}}\\
    \emph{$\decayFac \leftarrow \calDecayFac(\smoothingFac, \tuningFac)$}
    \Comment{Eq.~\ref{fuc:decay}}\\
    \emph{$\batchSize_\text{adapt}\leftarrow \lfloor m_\text{min}+\decayFac \cdot (m_\text{max}-m_\text{min})\rfloor$}\\
    \BlankLine
    \Return{$\batchSize_\text{adapt}$}
}

\end{algorithm}

\subsection{Non-linear Prolate Methods}\label{subsec: Prolate.}
\begin{algorithm}[t!]
\caption{\text{APT* - Calculate vertex charge}}
\label{alg: nonlinearProlateMethods}
\DontPrintSemicolon
% \scriptsize
\small
\SetKwInOut{Input}{Input}
\SetKwInOut{Output}{Output}
\SetKwFunction{normalize}{normalize}
\SetKwFunction{decayTanh}{tanhProlateMethod}
\Input{$\batchSize_\text{adapt}$ - The current adapted batch-size}
\Output{ $\charge_\text{i}$ - The suitable charge of the vertex}
\emph{$\charge_\text{min}\leftarrow 0.1, \charge_\text{max} \leftarrow 1.9, \charge_\text{i}\leftarrow \charge_\text{min}$}\\
\emph{$\epsilon \leftarrow 6, \beta \leftarrow -0.5$}\Comment{$\epsilon$ and $\beta$ are normalization constant}\\
$\batchSize_\text{biased}\leftarrow \epsilon\cdot(\normalize(\batchSize_\text{adapt})+\beta)$\\

$\charge_\text{i} \leftarrow \decayTanh(\batchSize_\text{biased},\charge_\text{min},\charge_\text{max})$ \Comment{Eq.~\ref{fuc:decayTanh}}\\

\Return $\charge_\text{i}$\\

\end{algorithm}
\begin{table}[t]
\caption{\textcolor{black}{Non-linear Prolation Methods Comparison (100 runs)}}
\centering
\vspace{-0.7em} 
\resizebox{0.485\textwidth}{!}{
\begin{tabular}{lccccccc}
 % \hline
 % \multicolumn{10}{|c|}{Contact Change Detection Accuracy} \\
 \toprule
 & \multicolumn{3}{c}{initial time} & \multicolumn{3}{c}{initial cost} &\multirow{2}*{\textbf{success}}\\
\cmidrule(lr){2-4} \cmidrule(lr){5-7}
    &$t^\textit{min}_\textit{init}$ &$t^\textit{med}_\textit{init}$ &$t^\textit{max}_\textit{init}$ &$c^\textit{min}_\textit{init}$ &$c^\textit{med}_\textit{init}$ &$c^\textit{max}_\textit{init}$ \\

\midrule
    $\textit{EIT*}$~\citeblue{strub2022adaptively}    &0.0490 &0.0768 &$\infty$ &1.6361 &2.4268 &$\infty$  &0.92\\
    $\textit{FIT*}$~\citeblue{Zhang2024adaptive}    &0.0484 &0.0688 &$\infty$ &1.5945 &2.1871 &$\infty$  &0.95\\
    $\textit{FDIT*}$~\citeblue{Zhang2024Elliptical}    &0.0494 &0.0658 &$\infty$ &1.5498 &1.9772 &$\infty$  &0.97\\
\midrule
    $\textit{APT*-E}$  &0.0473  &0.0650  &{0.1032} &1.5442 &1.9405 &2.7100 &{1.00}  \\
    $\textit{APT*-P}$  &0.0437  &{0.0647}  &$\infty$ &1.5148 &1.9259 &$\infty$  &0.97  \\
    $\textit{APT*-L}$  &0.0482  &0.0654  &0.1110 &1.5334 &{1.8941} &2.7410 &{1.00}  \\
    $\textit{APT*-I}$  &0.0472  &0.0657  &$\infty$ &1.6337 &1.9161 &$\infty$  &0.98  \\
 \midrule   
    \textcolor{black}{${\textit{{APT*-T}}}$ } \\ 
    
    \textcolor{black}{${\textit{$\alpha=10$}}$}   
    &{0.0439} &{0.0633} &{0.1075} &{1.5212} &{1.9120} &{2.2960}  &{1.00}\\
    
    \textcolor{black}{${\textit{$\alpha=100$}}$}  
    &{0.0441} &{0.0639} &{0.1056} &{1.4772} &{1.8826} &{2.2104}  &{1.00}\\
    
    \textcolor{black}{${\textit{$\alpha=1000$}}$}    
    &{0.0454} &{0.0648} &$\infty$ &{1.4657} &{1.8719} &$\infty$  &{0.99}\\
\bottomrule
%  success & \multicolumn{3}{|c}{0.48} & \multicolumn{3}{|c}{0.71} & \multicolumn{1}{|c}{\textbf{0.88}} \\
%  % power  & & & & & &\\
%  \hline
\end{tabular}} \label{tab: decay_method}
\vspace{-1.7em} 
\end{table}
\begin{figure}[t!]
    \centering
    \begin{tikzpicture}
    
    \node[inner sep=0pt] (russell) at (0.0,0.0)
    {\includegraphics[width=0.35\textwidth]{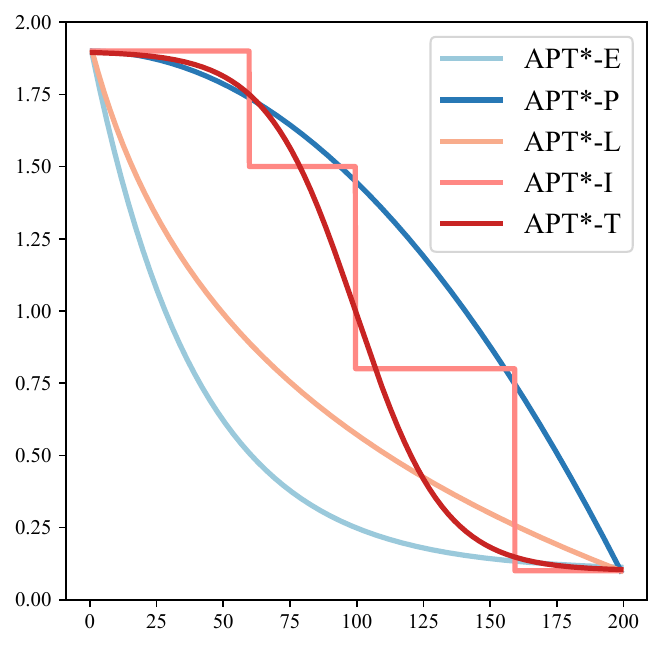}};

    \node [rotate=90] at (-3.4,0.0) {\small\textit{Vertices electric charge }$[q_i]$};
    
    \node at (0.0,-3.28) {\small\textit{Samples per batch }$\batchSize(\decayFac)$};
        
    \end{tikzpicture}
    \vspace{-1.2em} 
    \caption{This graph illustrates the comparison of the five non-linear prolate methods. The maximal electric charge is 1.9, while the minimal charge is 0.1. Similarly, the maximal batch size is 199, and the minimal batch size is 1.}
    \label{fig: decay_method}
    \vspace{-1.7em} 
\end{figure}
In this section, we evaluated five non-linear prolation methods based on the results from the force-based elliptical-RNN module (Sec.~\ref{subsec:elliptical RNN}) and dynamic batch-size tuning module (Sec.~\ref{subsec:adaptive-batch}).
\textcolor{black}{APT* adaptively adjusts the electric charge of each vertex based on the batch size, thereby influencing the magnitude of the force calculated using Coulomb's law. When the vertex charge increases, the resulting force intensifies, leading to a more pronounced hyperelliptical shape (i.e., higher eccentricity) of the neighbor region. Conversely, when the charge is set to zero, the force becomes null, and the neighbor region assumes a hypersphere shape (i.e., zero eccentricity).
}

As shown in Table \ref{tab: decay_method}, \( t^\textit{min}_\textit{init} \) denotes the minimal initial planning time, and \( c^\textit{max}_\textit{init} \) represents the maximal initial cost of the planning problem. For failed attempts, the \textit{cost} and \textit{time} metrics are denoted as infinity. Each method was evaluated across 100 independent runs to analyze its effectiveness in shaping the batch-size \batchSizeT.
Our comparative study investigated various non-linear prolate strategies to understand their distinctive impacts (Fig.~\ref{fig: decay_method}). \textcolor{black}{In Alg.~\ref{alg: nonlinearProlateMethods}, the scaling factors $\epsilon$ and $\beta$ facilitate the necessary transformation to align the corresponding function regarding the range of vertex charge $q$ and batch size \batchSizeT. The evaluation considered five approaches: \textit{Exponential} (APT*-E), \textit{Polynomial} (APT*-P), \textit{Logarithmic} (APT*-L), and \textit{Iteration-based} (APT*-I), \textit{Tanh} function (APT*-T).}

\begin{figure*}[t!]
    \centering
    \begin{tikzpicture}
    \node[anchor=center] at (0,0) 
    {\includegraphics[width=0.97\textwidth]{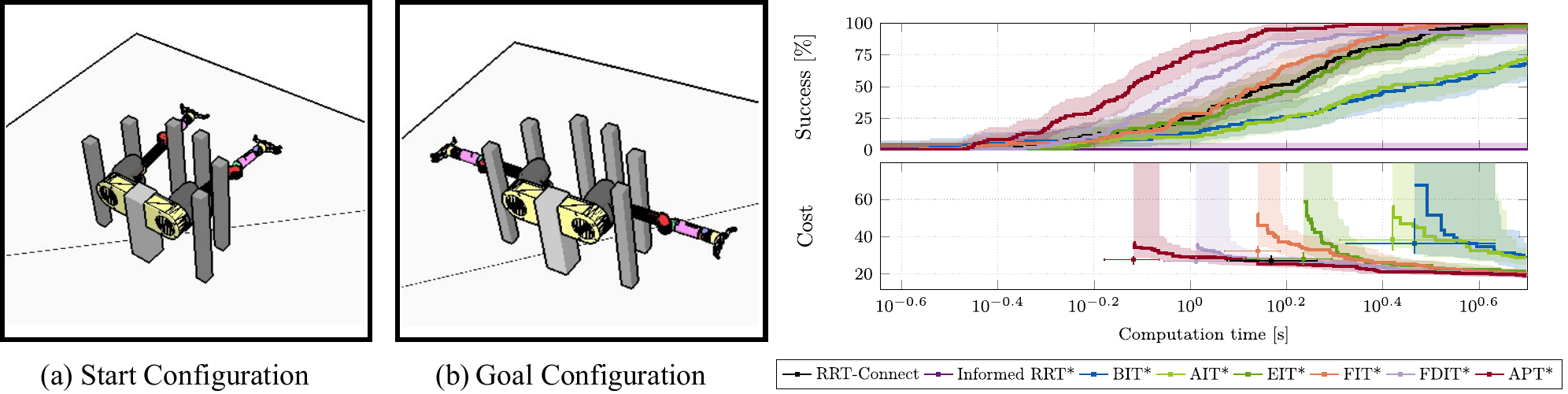}};
    \end{tikzpicture}
    \vspace{-1em}
    % \caption{Illustrations of the dual-arm manipulator ($\mathbb{R}^{14}$) problem in the \textit{cage}-ENV. The top row depicts the start configuration of the arms in an extended forward position within a constrained space (a–d). The bottom row presents the goal configuration, where the arms extend outward in opposite directions without colliding with the cage (e–h), with the results presented in Fig.~\ref{fig: sim_cage}.}
    \caption{\textcolor{black}{Illustrations of the dual-arm manipulator ($\mathbb{R}^{14}$) problem in the \textit{cage}-ENV. Fig.(a) depicts the start configuration of the arms in an extended forward position within a constrained space. Fig.(b) presents the goal configuration, where the arms extend outward in opposite directions without colliding with the cage. Detailed experimental results are presented above and in Table~\ref{tab:planner_comparison}. All planners have a maximum time of 5 seconds over 100 runs to solve this problem.}
}
    \label{fig: sim_cage}
    \vspace{-1.7em}
\end{figure*}
\begin{table*}[ht]

\caption{\textcolor{black}{Performance comparison in \text{cage}-ENV (Fig.~\ref{fig: sim_cage}) over 100 runs. Here, $t$ and $c$ denote time and cost; \texttt{init} and \texttt{final} refer to initial and final solutions; \texttt{min}, \texttt{med}, and \texttt{max} represent minimum, median, and maximum values. \textbf{Bold} indicates the best in each column. (Unsuccessful runs incur infinite time and cost.)}
}
    \centering
    \vspace{-1.0em} 
\resizebox{0.98\textwidth}{!}{
    \begin{tabular}{lcccccccccc}
        \toprule
        Planner & \(t_\mathrm{init}^\mathrm{min}\) & \(t_\mathrm{init}^\mathrm{med}\) & \(t_\mathrm{init}^\mathrm{max}\) & \(c_\mathrm{init}^\mathrm{min}\) & \(c_\mathrm{init}^\mathrm{med}\) & \(c_\mathrm{init}^\mathrm{max}\) & \(c_\mathrm{final}^\mathrm{min}\) & \(c_\mathrm{final}^\mathrm{med}\) & \(c_\mathrm{final}^\mathrm{max}\) & Success \\
        \midrule
        RRT-Connect~\citeblue{connect2000} & 0.4264 & 1.4739 & \(\infty\) & 13.1566 & 27.0866 & \(\infty\) & 13.1566 & 27.0866 & \(\infty\) & 0.98 \\
        Informed RRT*~\citeblue{gammell2014informed} & \(\infty\) & \(\infty\) & \(\infty\) & \(\infty\) & \(\infty\) & \(\infty\) & \(\infty\) & \(\infty\) & \(\infty\) & 0.00 \\
        BIT*~\citeblue{gammell2020batch} & \textbf{0.2270} & 2.9238 & \(\infty\) & 13.6402 & 36.2940 & \(\infty\) & 10.9949 & 28.6882 & \(\infty\) & 0.68 \\
        AIT*~\citeblue{strub2022adaptively} & 0.3836 & 2.6334 & \(\infty\) & 16.4872 & 38.3468 & \(\infty\) & 11.5357 & 28.6952 & \(\infty\) & 0.73 \\
        EIT*~\citeblue{strub2022adaptively} & 0.5533 & 1.7184 & \(\infty\) & 13.3572 & 28.1140 & \(\infty\) & 11.4210 & 20.8505 & \(\infty\) & 0.98 \\
        FIT*~\citeblue{Zhang2024adaptive} & 0.2749 & 1.3814 & 3.2350 & \textbf{12.0491} & 32.2499 & 68.6607 & 11.9310 & 20.3170 & 44.4034 & \textbf{1.00} \\
        FDIT*~\citeblue{Zhang2024Elliptical} & 0.3298 & 1.0828 & \(\infty\) & 12.1402 & \textbf{26.7375} & \(\infty\) & \textbf{10.7653} & \textbf{18.9521} & \(\infty\) & 0.93 \\
        APT* (ours) & 0.3377 & \textbf{0.7523} & \textbf{2.7564} & 13.3200 & 27.5476 & \textbf{43.4205} & 11.0353 & 19.1839 & \textbf{39.3813} & \textbf{1.00} \\
        \bottomrule
    \end{tabular}}
    
    \label{tab:planner_comparison}
    \vspace{-1.3em} 
\end{table*}
\begin{figure}[t!]
    \centering
    \begin{tikzpicture}
    \node[inner sep=0pt] (russell) at (-4.0,0.0)
    {\includegraphics[width=0.24\textwidth]{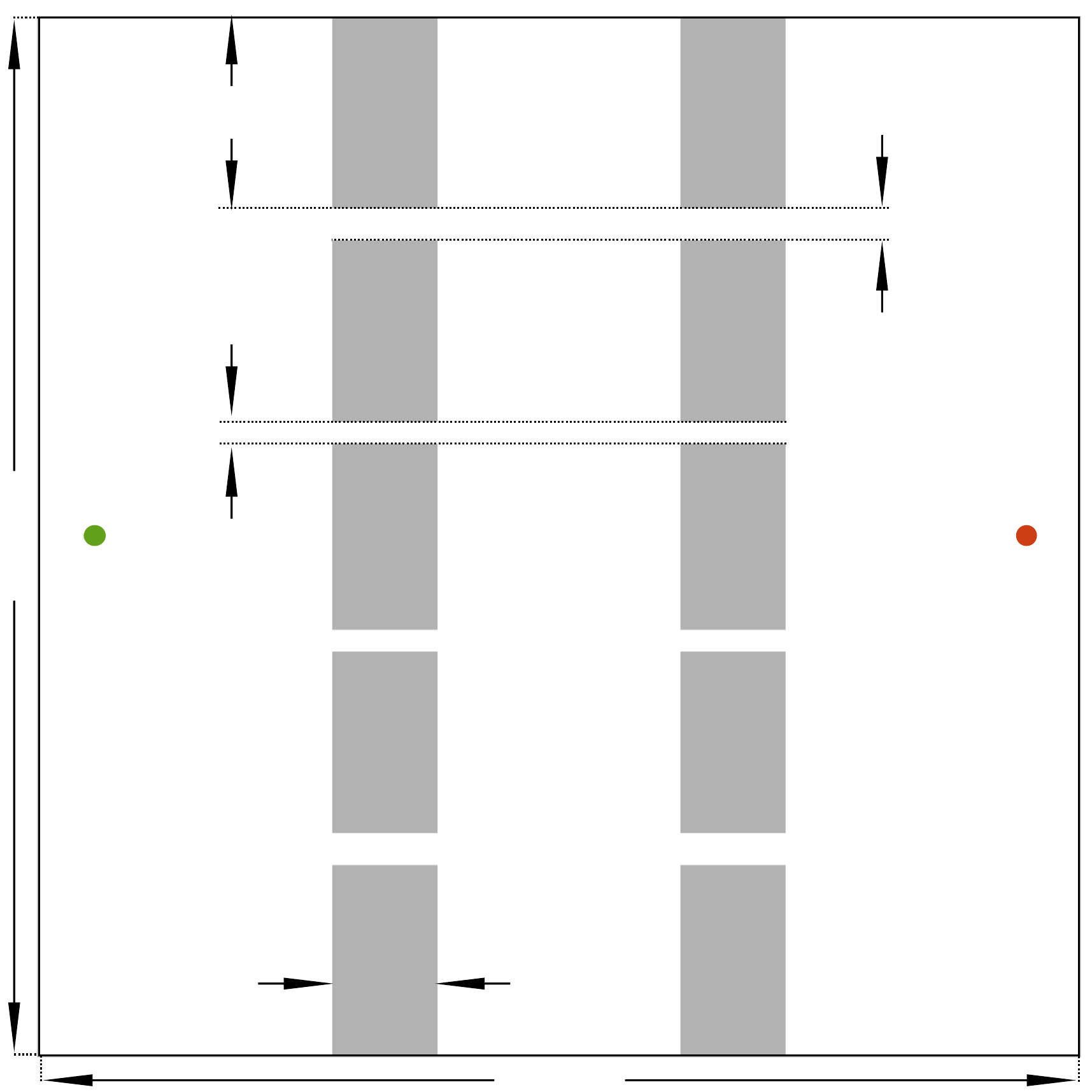}};
    \node[inner sep=0pt] (russell) at (0.25,0.0)
    {\includegraphics[width=0.24\textwidth]{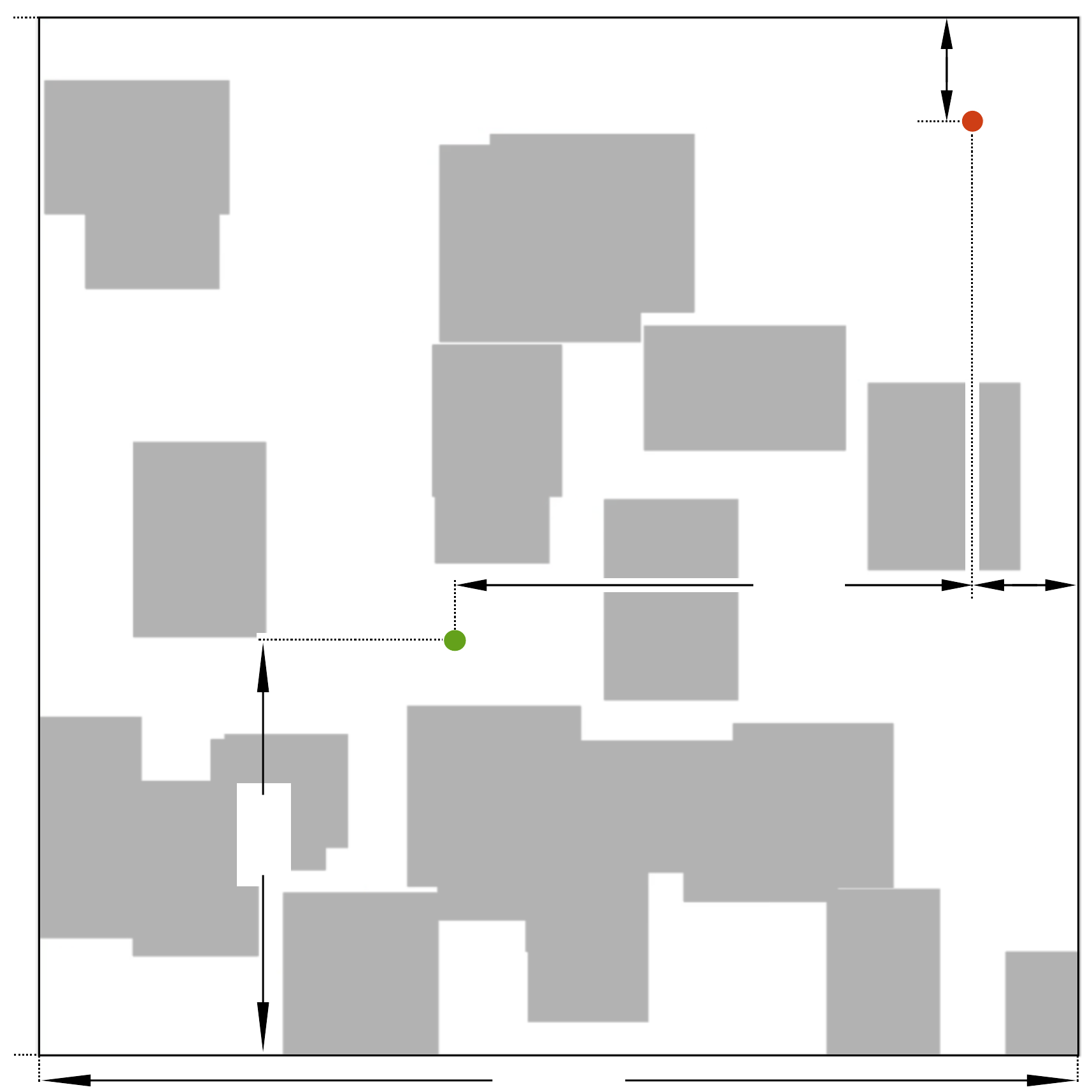}};
    \scriptsize
    
    %% wall Gap
    %%left
    % \node [rotate=90] at (-5.98,1.15) {0.5};
    \node [rotate=90] at (-5.45,0.5) {0.02};    
   
    \node [rotate=90] at (-5.45,1.7) {0.18};
    % \node at (-5.62,-0.46) {0.05};
    %%right

    \node [rotate=90] at (-2.5,1.3) {0.03};
    % \node [rotate=90] at (-1.92,-1.01) {0.5};
    % \node at (-2.12,0.55) {0.05};
    %%Down

    \node at (-4.2,-1.58) {0.1};

    %%frame
    \node [rotate=90] at (-6.12,0.05) {1.0};
    \node at (-4.0,-2.13) {1.0};
    \node at (-5.43,-0.5) {(0.05,0.5)};
    \node at (-2.47,-0.5) {(0.95,0.5)};
    \node at (-5.6,-0.2) {\color{teal} Start};
    \node at (-2.3,-0.2) {\color{purple} Goal};
    %% Random Rectangles
    \node at (1.24,-0.1) {0.5};
    \node at (0.25,-2.13) {1.0};
    \node [rotate=90] at (-0.89,-1.17) {0.4};
    \node [rotate=90] at (1.98,1.95) {0.1};
    \node  at (2.2,-0.3) {0.1};

    \node at (-0.1,-0.51) {\color{teal} Start};
    \node at (1.6,1.55) {\color{purple} Goal};

    \node at (-4.0,-2.51) {\small (a) Dividing Wall-gaps (DW)};
    \node at (0.25,-2.51) {\small (b) Random Rectangles (RR)};
    \end{tikzpicture}
    \vspace{-0.6em} 
    \caption{Simulated planning problems were visualized using a 2D representation. The possible states, represented as $X \subset \mathbb{R}^n$, were confined within a hypercube of side length 1.0 in both scenarios. Ten variations of the dividing wall gaps and random rectangles experiment were conducted, with the results presented in Fig.~\ref{fig: result}.}
    \label{fig: testEnv}
    \vspace{-1.2em} 
\end{figure}
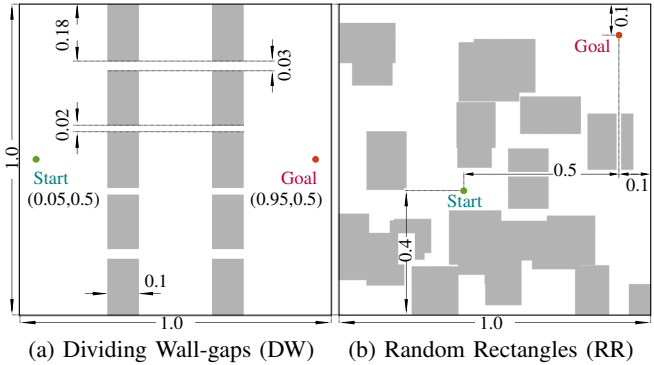

Experimental results from Table \ref{tab: decay_method} indicate that APT*-T consistently outperforms the other methods, achieving the lowest \textit{median time} and \textit{median cost} for initial solutions. The computational efficiency of the vertex charge in the APT*-T method is defined by the Taylor expansion of the \textit{Tanh} function, expressed as:  
\begin{equation}
\label{fuc:decayTanh}
    \begin{split}
    \charge (\batchSize)& := \frac{(\charge_\textit{min}+\charge_\textit{max})}{2}+\\&\sum_{i=1}^{\alpha}\frac{2^{2i-1}B_{2i}(2^{2i}-1){\batchSize(\decayFac)}^{2i-1}(\charge_\textit{min}-\charge_\textit{max})}{(2i)!},  
    \end{split}
\end{equation}
where $B_{2i} := \sum_{j=0}^{i}\sum_{k=0}^{j}\left(-1\right)^{k}\left(\begin{matrix}{j}\\{k}\\\end{matrix}\right)\frac{k^{2i}}{j+1}$ denotes as the $2i^{th}$ Bernoulli number and $\alpha$ determine the order of the Taylor series expansion. \textcolor{black}{As shown in Table~\ref{tab: decay_method}, a larger $\alpha$ (e.g., 1000) increases computation time but yields more accurate solutions. Conversely, a smaller $\alpha$ (e.g., 10) reduces pathfinding time at the cost of lower path quality.}
The electric charge of vertices, \( \charge (\batchSize) \), greatly optimizes the total magnitude of Coulomb force \compFT, acting on neighbors. 
This force plays a crucial role in adjusting the eccentricity of elliptical-RNN. 
\textcolor{black}{The \( n \)-th order eccentricity of the hyperellipsoid is defined as:
\begin{equation}
e_n := 0= \lim_{\compF(\charge) \to 0}\sqrt{1 - \frac{r(\batchSize)}{\left(\prod_{i=1}^n d_i\right)^{1/n}}},
\end{equation}
where the eccentricity \(e_n\) measure uses the normalized geometric mean. When the electric charge \(\charge\) approaches \(\charge_{\textit{min}}\) (i.e., \(\compF \rightarrow 0\)), the length of the stretching axis \(d_{1}\) approaches neighbor radius \(r(\batchSize)\), thereby \(e_n\) of the elliptical-RNN approaches zero, corresponding to a hypersphere.} 

\section{Experiments}\label{sec:Expri}

% %
% \begin{figure}[t!]
%     \centering
%     \begin{tikzpicture}

%     \node[inner sep=0pt] (russell) at (0,0)
%     {\includegraphics[width=0.48\textwidth]{figure/benchmark/cage_8.pdf}};

%      \node[inner sep=0pt] (russell) at (0.48,-2.5)
%     {\includegraphics[width=0.4\textwidth]{figure/benchmark/bar8_test.pdf}};
%     \end{tikzpicture}
%     \vspace{-0.6em} 
%     \caption{\textcolor{purple}{Detailed experimental results from \textit{cage}-ENV are presented above. All planners have a maximum time of 5 seconds over 100 runs to solve this problem. APT* results in the fastest convergence of the optimal solution.}}

%     \label{fig: sim_cage}
%     \vspace{-1.7em} 
% \end{figure}
% 
%
\begin{figure*}[t!]
    \centering
    \begin{tikzpicture}
    \node[inner sep=0pt] (russell) at (4.1,8)
    {\includegraphics[width=0.49\textwidth]{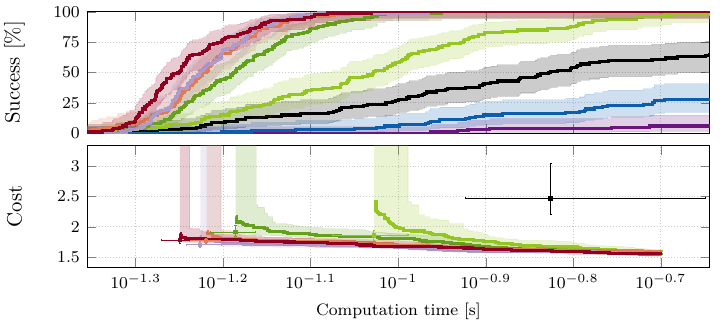}};
    \node[inner sep=0pt] (russell) at (4.1,3.5)
    {\includegraphics[width=0.49\textwidth]{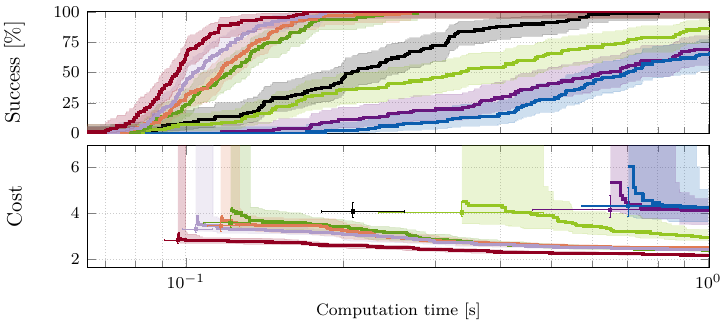}};
    \node[inner sep=0pt] (russell) at (4.1,-1)
    {\includegraphics[width=0.49\textwidth]{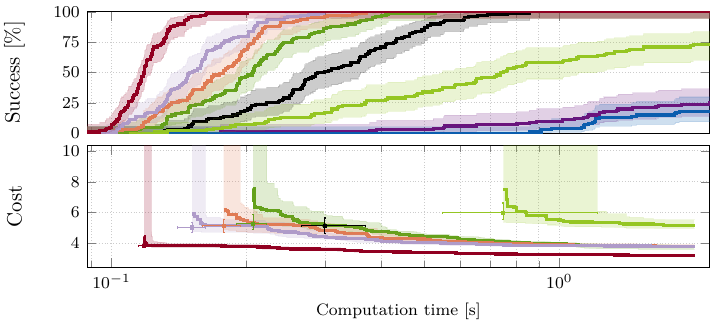}};

    \node[inner sep=0pt] (russell) at (-4.9,8)
    {\includegraphics[width=0.49\textwidth]{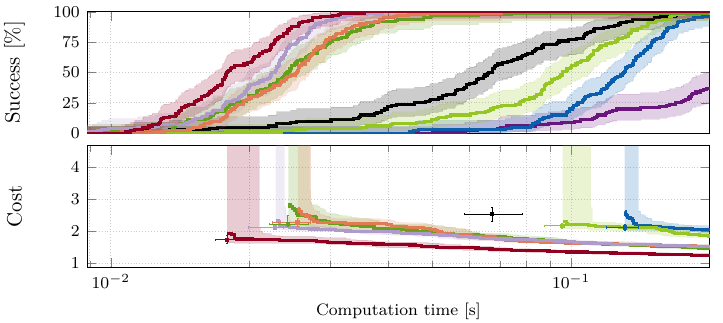}};  
    \node[inner sep=0pt] (russell) at (-4.9,3.5)
    {\includegraphics[width=0.49\textwidth]{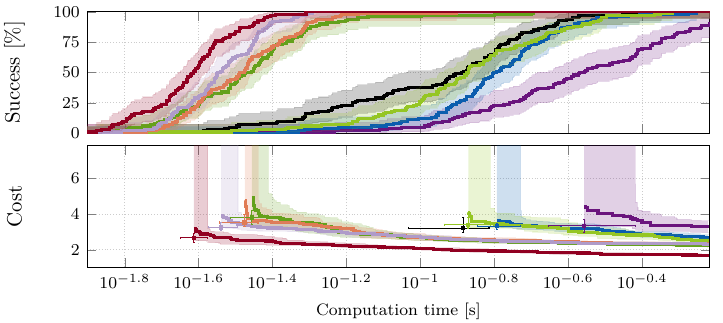}};
    \node[inner sep=0pt] (russell) at (-4.9,-1){\includegraphics[width=0.49\textwidth]{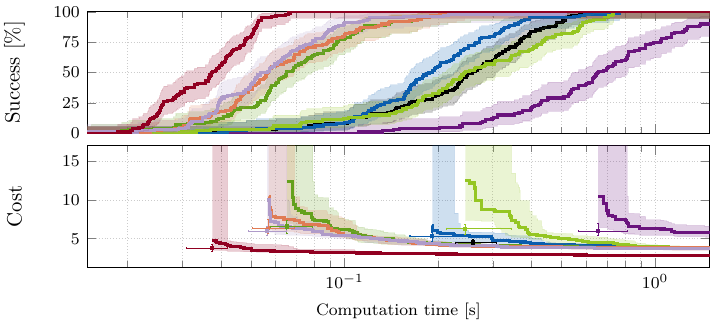}};

    \node[inner sep=0pt] (russell) at (0.0,-3.9){\includegraphics[width=0.8\textwidth]{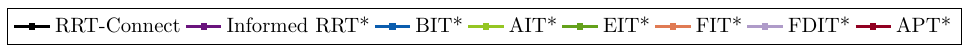}};

    \node at (-4.5,5.8) {\footnotesize (a) Dividing Wall-gaps (DW) in $\mathbb{R}^4$ - MaxTime: 0.2s};
    \node at (-4.5,1.3) {\footnotesize (c) Dividing Wall-gaps (DW) in $\mathbb{R}^8$ - MaxTime: 0.6s};
    \node at (-4.5,-3.2) {\footnotesize(e) Dividing Wall-gaps (DW) in $\mathbb{R}^{16}$ - MaxTime: 1.5s};

    \node at (4.5,5.8) {\footnotesize (b) Random Rectangles (RR) in $\mathbb{R}^4$ - MaxTime: 0.2s};
    \node at (4.5,1.3) {\footnotesize(d) Random Rectangles (RR) in $\mathbb{R}^8$ - MaxTime: 1.0s};
    \node at (4.5,-3.2) {\footnotesize(f) Random Rectangles (RR) in $\mathbb{R}^{16}$ - MaxTime: 2.0s};

    \end{tikzpicture}
    \vspace{-1.0em} 
    \caption{\textcolor{black}{Detailed experimental results from Section~\ref{subsec:experi} are presented above. MaxTime is the planner's maximum allotted planning time. Fig. (a), (c), and (e) depict test benchmark dividing walls (DW) outcomes in $\mathbb{R}^4$ to $\mathbb{R}^{16}$, respectively. Panel (b) showcases random rectangle (RR) experiments in $\mathbb{R}^4$, while panels (d) and (f) demonstrate in $\mathbb{R}^8$ and $\mathbb{R}^{16}$. In the cost plots, boxes represent solution cost and time, with lines showing cost progression for optimal planners (unsuccessful runs have infinite costs). Error bars provide nonparametric 99\% confidence intervals for solution cost and time.}}
    \label{fig: result}
    \vspace{-1.8em}
\end{figure*}
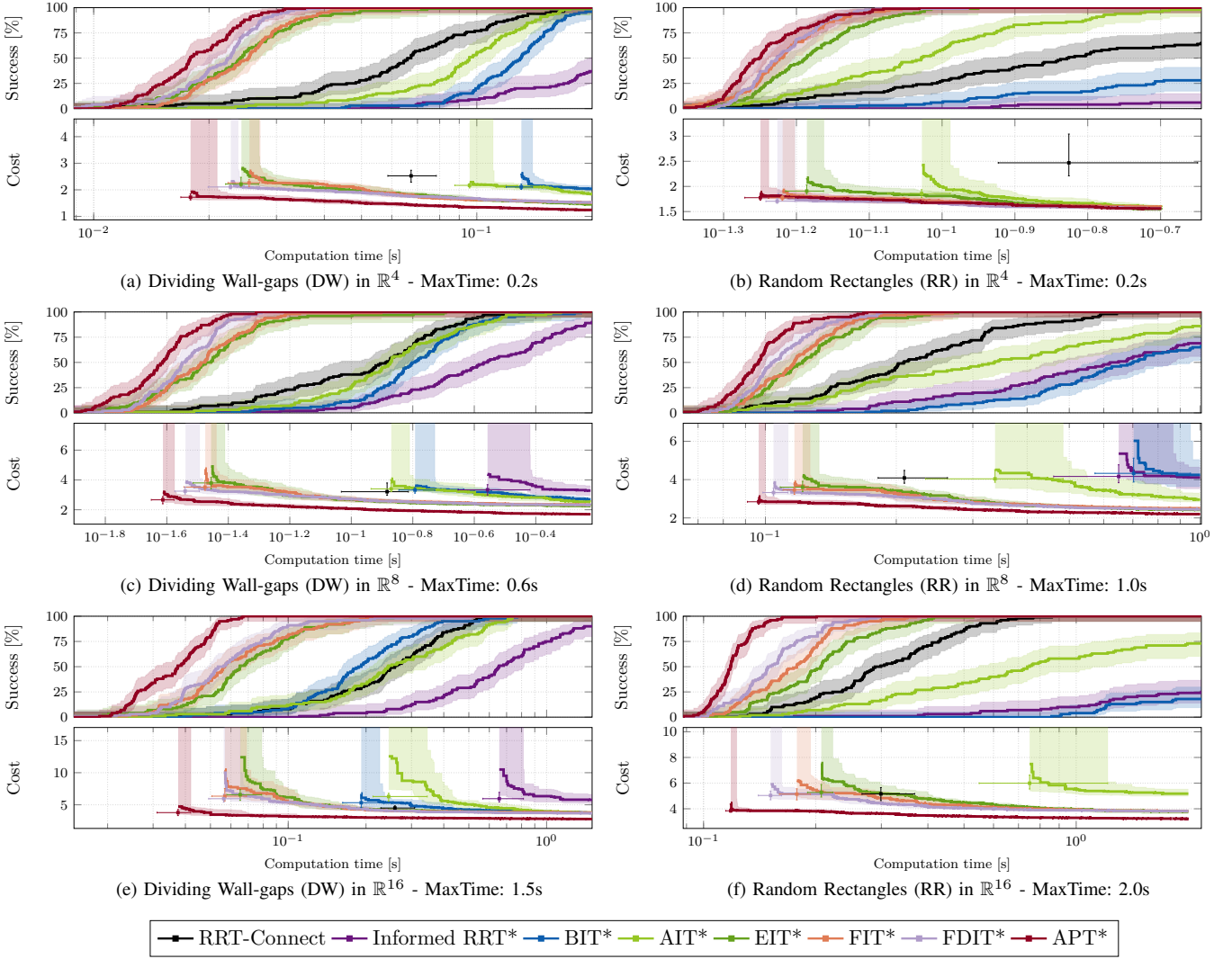
\begin{figure*}[t!]
    \centering
    \begin{tikzpicture}
    \node[inner sep=0pt] (russell) at (0,0)
    {\includegraphics[width=0.95\textwidth]{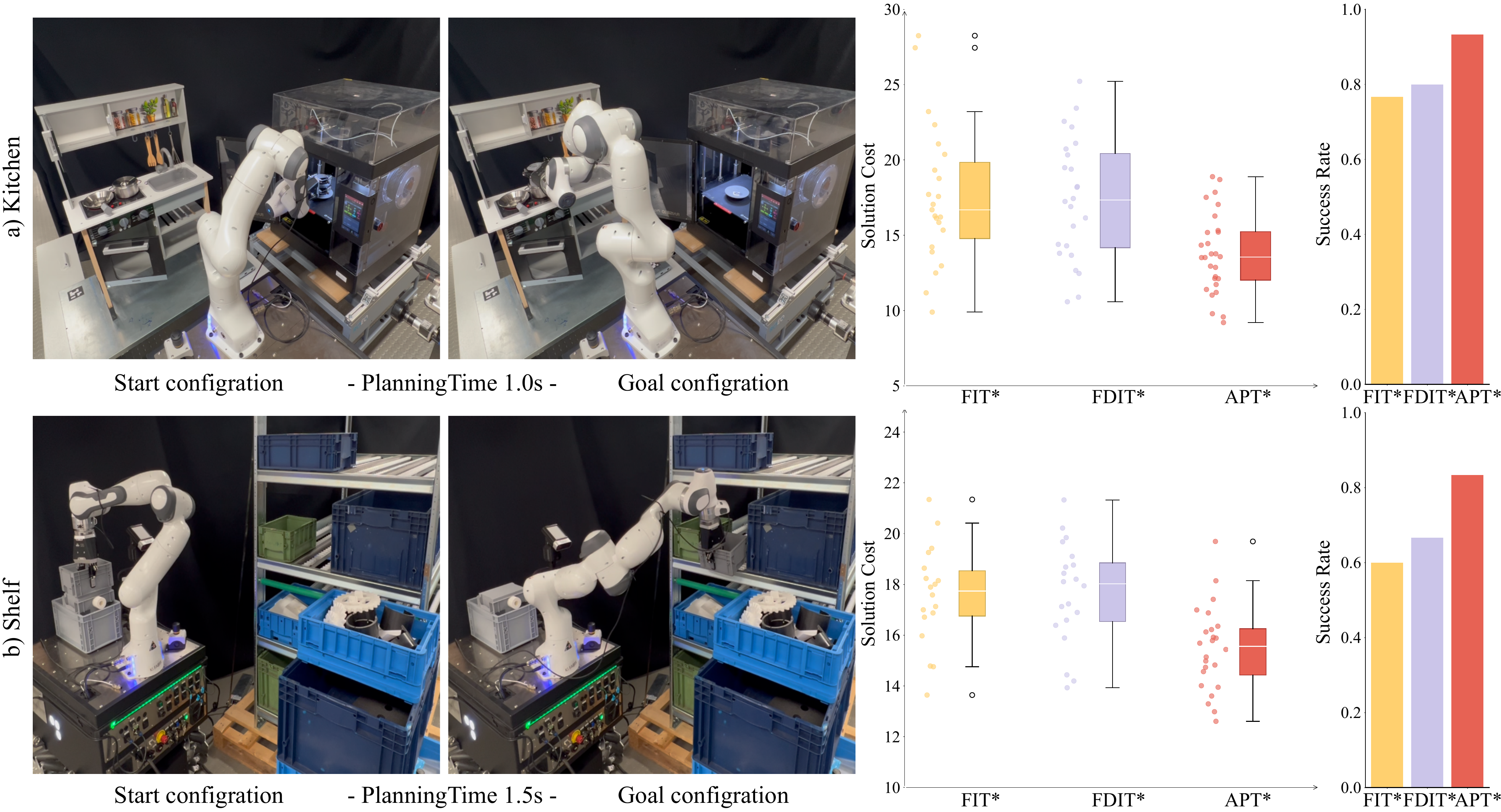}};

    \end{tikzpicture}
    \vspace{-0.3em} 
    \caption{\textcolor{black}{Experimental results from Section~\ref{subsec:realExpri} are summarized above. Fig.~\ref{fig: Realresult}a shows the \textit{kitchen}-ENV with a manipulator taking printed parts from the printer to the kitchen model. Fig.~\ref{fig: Realresult}b highlights the \textit{shelf}-ENV, showing the start/goal configurations for handling an industry-standard (tolerance $\pm$5mm) container. Cost box plots display solution costs per planner, with white lines indicating mean cost progression (unsuccessful runs assigned infinite cost).}}
    \label{fig: Realresult}
    \vspace{-1.7em}
\end{figure*}

In this paper, we utilize the Planner-Arena benchmark database~\citeblue{moll2015benchmarking}, the Planner Developer Tools (PDT)~\citeblue{gammell2022planner}, and MoveIt~\citeblue{gorner2019moveit} to benchmark motion planner behaviors. APT* was evaluated against popular algorithms in both simulated random scenarios (Fig.~\ref{fig: testEnv}), path planning problems for dual-Barrett Whole-Arm Manipulator (dual-WAM-$\mathbb{R}^{14}$) in the Open Robotics Automation Virtual Environment (OpenRAVE, Fig.~\ref{fig: sim_cage})~\citeblue{Diankov2008OpenRAVEAP} and real-world manipulation problems (Fig.~\ref{fig: Realresult}). The comparison involved RRT-Connect, Informed RRT*, BIT*, AIT*, EIT*, FIT*, and FDIT* sourced from the Open Motion Planning Library (OMPL)~\citeblue{sucan2012open}. 
% The evaluation environment is with an Intel i7 3.90 GHz processor and 32GB of LPDDR3 3200 MHz memory. 
These comparisons were tested in simulated tasks of dimensions $\mathbb{R}^4$ to $\mathbb{R}^{16}$. The primary objective for all the planners was to minimize path length (cost). For RRT-based algorithms, a goal bias of 5\% was incorporated, with maximum edge lengths of 0.5, 1.25, and 3.0 in $\mathbb{R}^4$, $\mathbb{R}^8$, $\mathbb{R}^{16}$. The RGG constant $\eta$ was uniformly set to 1.001, and the rewire factor was set to 1.2 for all planners. \textcolor{black}{The implementation of APT* planner into OMPL framework is available at: }\href{https://github.com/Liding-Zhang/ompl_apt.git}{\textcolor{black}{https://github.com/Liding-Zhang/ompl\_apt.git}}
%
 
% All batched algorithms other than FIT* and APT* utilized a batch size of 100. All planners employed the Euclidean distance and effort as a heuristic, respectively. 

% APT*'s adaptively prolate search regions according to current batch-size. The planner's adaptive mechanism optimizes the neighbor search areas and rewires at any time.
%
\subsection{Simulation Experimental Tasks}\label{subsec:experi}
\begin{table*}[t]
\caption{\textcolor{black}{Benchmark Evaluations ($\Uparrow$ and color represent the corresponding medium initial time improvement) (Fig.~\ref{fig: result} and Fig.~\ref{fig: sim_cage})}}
\centering
\vspace{-1.0em} 
\resizebox{0.92\textwidth}{!}{
\begin{tabular}{lcccccccccc}
 % \hline
 % \multicolumn{10}{|c|}{Contact Change Detection Accuracy} \\
\toprule
& \multicolumn{3}{c}{${\text{Flexible Informed Trees}}$}& \multicolumn{3}{c}{${\text{Force Direction Informed Trees}}$} & \multicolumn{3}{c}{${\textcolor{black}{\text{Adaptively Prolated Trees (ours)}}}$} &\multirow{2}*{\normalsize$t^\textit{med}_\textit{init}\color{violet}\Uparrow \color{black}{/} \color{teal}\Uparrow$ (\%)}\\
\cmidrule(lr){2-4} \cmidrule(lr){5-7} \cmidrule(lr){8-10}
%  \cline{2-7}
    &$t^\textit{med}_\textit{init}$ &$c^\textit{med}_\textit{init}$ &$c^\textit{med}_\textit{final}$ &$t^\textit{med}_\textit{init}$ &$c^\textit{med}_\textit{init}$ &$c^\textit{med}_\textit{final}$ 
    &$t^\textit{med}_\textit{init}$ &$c^\textit{med}_\textit{init}$ &$c^\textit{med}_\textit{final}$\\
  \midrule 
    $\text{DW}-\mathbb{R}^4$   &\textcolor{violet}{0.0252}   &2.2646   &1.5143 &\textcolor{teal}{0.0308} &2.1067 &1.5268 &\textcolor{purple}{0.0179} &1.7210 &1.2394 &\textcolor{violet}{{28.96}} / \textcolor{teal}{21.49} \\
    $\text{DW}-\mathbb{R}^8$   &\textcolor{violet}{0.0334}   &3.5209   &2.3365 &\textcolor{teal}{0.0288} &{3.2466} &2.3093  &\textcolor{purple}{0.0244} &{2.6658} &1.7006 &\textcolor{violet}{{26.94}} / \textcolor{teal}{20.77}\\
    $\text{DW}-\mathbb{R}^{16}$   
    &\textcolor{violet}{0.0568} &{6.3452} &{3.7953} &\textcolor{teal}{0.0563} &{5.9502} &{3.6954}  &{\textcolor{purple}{0.0374}} &{3.7873} &{2.7854} &\textcolor{violet}{34.15} / \textcolor{teal}{33.57}\\
  \midrule   
    $\text{RR}-\mathbb{R}^4$    &\textcolor{violet}{0.0603} &1.7762 &1.5929 &\textcolor{teal}{0.0594} &1.7050 &1.5616 &{\textcolor{purple}{0.0538}} &1.7725 &1.5597  &\textcolor{violet}{\text{10.77}} / \textcolor{teal}{9.43}\\
    $\text{RR}-\mathbb{R}^8$  &\textcolor{violet}{0.1166}  &3.4637  &2.5049 &\textcolor{teal}{0.1078} &3.3235  &2.4449
    &{\textcolor{purple}{0.0949}} &2.8425 &{2.1976} &\textcolor{violet}{18.61} / \textcolor{teal}{11.96}\\
    $\text{RR}-\mathbb{R}^{16}$ &\textcolor{violet}{0.1780}  &5.1412  &3.8146 &\textcolor{teal}{0.1513}  &5.0367  &3.8190 &{\textcolor{purple}{0.1183}} &{3.8367} &{3.2146}  &\textcolor{violet}{{33.54}} / \textcolor{teal}{\text{21.81}}\\
   \midrule  
    $\text{Cage}-\mathbb{R}^{14}$ &\textcolor{violet}{1.3814}   &32.2499   &20.3170 &\textcolor{teal}{1.0828}   &26.7375   &19.1839 &{\textcolor{purple}{0.7523}} &{27.5476} &{18.9521}  &\textcolor{violet}{{45.54}} / \textcolor{teal}{\text{30.52}}  \\
\bottomrule
%  success & \multicolumn{3}{|c}{0.48} & \multicolumn{3}{|c}{0.71} & \multicolumn{1}{|c}{\textbf{0.88}} \\
%  % power  & & & & & &\\
%  \hline
\end{tabular}} \label{tab:benchmark}
\vspace{-1.3em} 
\end{table*}
The planners were evaluated across three distinct benchmarks in three domains: $\mathbb{R}^4$, $\mathbb{R}^8$, and $\mathbb{R}^{16}$. In the first scenario, a constrained environment resembling a dividing wall (DW) with several narrow gaps was simulated, allowing valid paths in multiple general directions for non-intersecting solutions (Fig.~\ref{fig: testEnv}a). Each planner was tested over 100 runs, with computation times for each anytime asymptotically optimal planner detailed in the labels, using varying random seeds. The overall success rates and median path lengths for all planners are presented in Fig.~\ref{fig: result}a, Fig.~\ref{fig: result}c, and Fig.~\ref{fig: result}e. Results indicate that APT* quickly finds an initial solution in all dimensions with minimal time, whereas other planners require more time.

In the second test scenario, random widths were assigned to \textit{axis-aligned hyperrectangles}, generated arbitrarily within the \textit{$\mathcal{C}$-space} (Fig.~\ref{fig: testEnv}b). Random rectangle (RR) problems were created for each dimension of the \textit{$\mathcal{C}$-space}, with each planner undergoing 100 runs for every instance. Fig.~\ref{fig: result}b, Fig.~\ref{fig: result}d, and Fig.~\ref{fig: result}f illustrate the proposed method has the highest success rates and lowest median path costs within the computation time compared with other planners. 
% This indicates that APT* can explore via prolated search regions, thereby biasing the sampling process toward elliptical-RNN. 
% As a result, APT* outperformed and can quickly find an initial solution.

In the final test scenario, we evaluated APT* in the dual-WAM \textit{cage}-ENV for handling intrinsic settings (Fig.~\ref{fig: sim_cage}). The \textit{cage} problems were conducted in $\mathbb{R}^{14}$, with each planner tested over 100 runs in 5 seconds for every instance. \textcolor{black}{Table~\ref{tab:planner_comparison} illustrates that the adaptively prolated approach achieved the highest initial convergence rate and fastest computation time.}

As observed in Table~\ref{tab:benchmark}, there is a noticeable improvement in median initial times across varied benchmark scenarios, correlating with dimensionality. For instance, in the $\text{DW}-\mathbb{R}^4$ scenario, APT* demonstrates a faster initial median time (i.e., the median value over 100 trials) of maximal 28.96\% compared to FDIT* and FIT*. This trend persists across other scenarios, such as $\text{RR}-\mathbb{R}^8$ and $\text{Cage}-\mathbb{R}^{14}$, where APT* consistently convergence faster, including an approximately 30.52\% improvement in the \textit{cage} scenario compared to FDIT*.

Overall, APT* achieves superior performance compared to other motion planners. This improvement is attributed to its use of adaptively prolated elliptical-RNN search regions.
\subsection{Real-world Path Planning Tasks}\label{subsec:realExpri}
% %
To evaluate the algorithm's performance in real-world scenarios, two numerical experiments are conducted on a base-manipulator (DARKO) platform to demonstrate the efficiency and extensibility of the proposed path planner. 
We evaluate APT* alongside optimal path planners FIT* and FDIT*, comparing their performance in converging to the optimal solution cost and success rate over 30 runs. A collision-free path connecting the start state to the goal region is required. APT* demonstrated its effective prolation method during The \textit{kitchen printer}-ENV features manipulate 3D-printed kitchen tools, while the \textit{Shelf}-ENV environment involves navigating through industry standard narrow spaces (Fig.~\ref{fig: Realresult}). 
\subsubsection{\textbf{Kitchen Model Tool Printing Task}}
In the first task, we utilized the DARKO robot positioned in front of the kitchen and the 3D printer model. The start and goal configurations are illustrated in Fig.~\ref{fig: Realresult}a. This task is particularly challenging as the manipulator must navigate the geometric shape of the printed tool within a cluttered space while also avoiding collisions between the base robot, printer wall, and the kitchen shelves. The complexity is further heightened by the need for precise movements in a confined space.
Each planner was allotted 1.0 seconds to solve this kitchen tool print reallocation problem. Over the course of 30 trials, FIT* managed a 76.66\% success rate with a median solution cost of 16.2743. FDIT* had a success rate of 80\% with a median solution cost of 16.7816. APT* has the highest success rate of 93.33\% and the lowest average solution cost of 13.6948.
\subsubsection{\textbf{Industry Shelf Container Rearrangement Task}}
The initial and final configurations for the shelf task are depicted in Fig.~\ref{fig: Realresult}b. This task involves carrying an industry-standard container from the base robot and repositioning it on the third shelf layer between two containers. Due to component standardization (tolerance $\leq$ 5mm), the challenge lies in precisely inserting industry containers into narrow spaces. The task aims to place the industry-standard container between two larger containers on the shelf, making the planning of a collision-free feasible path particularly difficult.
Each planner was allocated 1.5 seconds to solve this confined, limited space carry-on and insertion problem. Across 30 trials, FIT* managed a 60\% success rate with a median solution cost of 17.7357. FDIT* had a 66.67\% success rate with a median solution cost of 18.0254. APT* achieved the highest success rate of 83.33\% with the best median solution cost of 15.7516.

In short, compared with the FIT* and the FDIT*, the APT* achieves the best performance in finding the initial solution and converging to the optimal solution.

\section{Conclusion and Future Work}
\textcolor{black}{In this paper, we introduce the Adaptively Prolated Trees (APT*), an extension of the Force Direction Informed Trees (FDIT*)~\citeblue{Zhang2024Elliptical} that employs a non-linear prolation-based nearest neighbor approach.} APT* assigns electric charge properties to vertices and utilizes an adaptive batch-size module to optimize them. These charges determine the Coulomb force magnitude via Coulomb's law, adjusting the prolated distance for elliptical $r$-nearest neighbors. 
APT* optimizes the prolated distances to prevent premature convergence to local minima and adjusts the nearest explore region during path optimization.
% Simulations across dimensions and real-world manipulation tasks validate APT*'s faster initial convergence and shorter path lengths than tested benchmarks.
% The real-world tasks further substantiate the practical efficacy of our approach, showcasing its robustness in real-world applications.

\textcolor{black}{Future research could focus on integrating human awareness and safety constraints into APT* for planning in dynamic scenarios (i.e., via local motion planners) and utilizing single instruction/multiple data parallelism methods~\citeblue{Wilson2024fcit} to reduce computation effort, enhancing overall planning efficiency.}

\bibliographystyle{IEEEtran}
\bibliography{bibliography}

\begin{thebibliography}{10}
\providecommand{\url}[1]{#1}
\csname url@samestyle\endcsname
\providecommand{\newblock}{\relax}
\providecommand{\bibinfo}[2]{#2}
\providecommand{\BIBentrySTDinterwordspacing}{\spaceskip=0pt\relax}
\providecommand{\BIBentryALTinterwordstretchfactor}{4}
\providecommand{\BIBentryALTinterwordspacing}{\spaceskip=\fontdimen2\font plus
\BIBentryALTinterwordstretchfactor\fontdimen3\font minus \fontdimen4\font\relax}
\providecommand{\BIBforeignlanguage}[2]{{%
\expandafter\ifx\csname l@#1\endcsname\relax
\typeout{** WARNING: IEEEtran.bst: No hyphenation pattern has been}%
\typeout{** loaded for the language `#1'. Using the pattern for}%
\typeout{** the default language instead.}%
\else
\language=\csname l@#1\endcsname
\fi
#2}}
\providecommand{\BIBdecl}{\relax}
\BIBdecl

\bibitem{gammell2021asymptotically}
J.~D. Gammell and M.~P. Strub, ``Asymptotically optimal sampling-based motion planning methods,'' \emph{Annual Review of Control, Robotics, and Autonomous Systems}, vol.~4, pp. 295--318, 2021.

\bibitem{Orthey2023AnnualReview}
A.~Orthey, C.~Chamzas, and L.~E. Kavraki, ``Sampling-based motion planning: A comparative review,'' \emph{Annual Review of Control, Robotics, and Autonomous Systems}, vol.~7, no.~1, pp. 285 -- 310, 2024.

\bibitem{zhang2024review}
L.~Zhang, K.~Cai, Z.~Sun, Z.~Bing, C.~Wang, L.~Figueredo, S.~Haddadin, and A.~Knoll, ``Motion planning for robotics: A review for sampling-based planners,'' \emph{Biomimetic Intelligence and Robotics}, vol.~5, no.~1, p. 100207, 2025.

\bibitem{penrose2003random}
M.~Penrose, \emph{Random geometric graphs}.\hskip 1em plus 0.5em minus 0.4em\relax OUP Oxford, 2003, vol.~5.

\bibitem{zhang2025TASE}
L.~Zhang, K.~Cai, Y.~Zhang, Z.~Bing, C.~Wang, F.~Wu, S.~Haddadin, and A.~Knoll, ``Estimated informed anytime search for sampling-based planning via adaptive sampler,'' \emph{IEEE Transactions on Automation Science and Engineering}, vol.~22, pp. 18\,580--18\,593, 2025.

\bibitem{dijkstra1959note}
E.~Dijkstra, ``A note on two problems in connexion with graphs,'' \emph{Numerische Mathematik}, vol.~1, no.~1, pp. 269--271, 1959.

\bibitem{hart1968formal}
P.~E. Hart, N.~J. Nilsson, and B.~Raphael, ``A formal basis for the heuristic determination of minimum cost paths,'' \emph{IEEE transactions on Systems Science and Cybernetics}, vol.~4, no.~2, pp. 100--107, 1968.

\bibitem{lavalle1998rapidly}
S.~LaValle, ``Rapidly-exploring random trees: A new tool for path planning,'' \emph{Research Report 9811}, 1998.

\bibitem{kavraki1996probabilistic}
L.~E. Kavraki, P.~Svestka, J.-C. Latombe, and M.~H. Overmars, ``Probabilistic roadmaps for path planning in high-dimensional configuration spaces,'' \emph{IEEE Transactions on Robotics and Automation}, vol.~12, no.~4, pp. 566--580, 1996.

\bibitem{connect2000}
J.~Kuffner and S.~LaValle, ``{RRT}-connect: An efficient approach to single-query path planning,'' in \emph{Proceedings 2000 ICRA. Millennium Conference. IEEE International Conference on Robotics and Automation. Symposia Proceedings (Cat. No.00CH37065)}, vol.~2, 2000, pp. 995--1001 vol.2.

\bibitem{karaman2011sampling}
S.~Karaman and E.~Frazzoli, ``Sampling-based algorithms for optimal motion planning,'' \emph{The international journal of robotics research}, vol.~30, no.~7, pp. 846--894, 2011.

\bibitem{fmt2015}
L.~Janson, E.~Schmerling, A.~Clark, and M.~Pavone, ``Fast marching tree: A fast marching sampling-based method for optimal motion planning in many dimensions,'' \emph{The International Journal of Robotics Research}, vol.~34, no.~7, pp. 883--921, 2015.

\bibitem{Zhang2024Elliptical}
L.~Zhang, Z.~Bing, Y.~Zhang, K.~Cai, L.~Chen, F.~Wu, S.~Haddadin, and A.~Knoll, ``Elliptical k-nearest neighbors: Path optimization via coulomb's law and invalid vertices in c-space obstacles,'' \emph{2024 IEEE/RSJ International Conference on Intelligent Robots and Systems (IROS)}, pp. 12\,032--12\,039, 2024.

\bibitem{gammell2014informed}
J.~D. Gammell, S.~S. Srinivasa, and T.~D. Barfoot, ``Informed {RRT}*: Optimal sampling-based path planning focused via direct sampling of an admissible ellipsoidal heuristic,'' in \emph{2014 IEEE/RSJ international conference on intelligent robots and systems}.\hskip 1em plus 0.5em minus 0.4em\relax IEEE, 2014.

\bibitem{gammell2018informed}
J.~D. Gammell, T.~D. Barfoot, and S.~S. Srinivasa, ``Informed sampling for asymptotically optimal path planning,'' \emph{IEEE Transactions on Robotics}, vol.~34, no.~4, pp. 966--984, 2018.

\bibitem{gammell2020batch}
J.~D. \hspace{0em}Gammell, T.~D. Barfoot, and S.~S. Srinivasa, ``Batch informed trees ({BIT}*): Informed asymptotically optimal anytime search,'' \emph{The International Journal of Robotics Research}, vol.~39, no.~5, 2020.

\bibitem{strub2022adaptively}
M.~P. Strub and J.~D. Gammell, ``Adaptively informed trees ({AIT}*) and effort informed trees ({EIT}*): Asymmetric bidirectional sampling-based path planning,'' \emph{The International Journal of Robotics Research}, vol.~41, no.~4, pp. 390--417, 2022.

\bibitem{Zhang2024adaptive}
L.~Zhang, Z.~Bing, K.~Chen, L.~Chen, K.~Cai, Y.~Zhang, F.~Wu, P.~Krumbholz, Z.~Yuan, S.~Haddadin, and A.~Knoll, ``Flexible informed trees ({FIT}*): Adaptive batch-size approach in informed sampling-based path planning,'' \emph{2024 IEEE/RSJ International Conference on Intelligent Robots and Systems (IROS)}, pp. 3146--3152, 2024.

\bibitem{zhang25ral}
L.~Zhang, Y.~Ling, Z.~Bing, F.~Wu, S.~Haddadin, and A.~Knoll, ``Tree-based grafting approach for bidirectional motion planning with local subsets optimization,'' \emph{IEEE Robotics and Automation Letters}, vol.~10, no.~6, pp. 5815--5822, 2025.

\bibitem{solovey2020critical}
K.~Solovey and M.~Kleinbort, ``The critical radius in sampling-based motion planning,'' \emph{The International Journal of Robotics Research}, vol.~39, no. 2-3, pp. 266--285, 2020.

\bibitem{moll2015benchmarking}
M.~Moll, I.~A. Sucan, and L.~E. Kavraki, ``Benchmarking motion planning algorithms: An extensible infrastructure for analysis and visualization,'' \emph{IEEE Robotics \& Automation Magazine}, vol.~22, no.~3, pp. 96--102, 2015.

\bibitem{gammell2022planner}
J.~D. Gammell, M.~P. Strub, and V.~N. Hartmann, ``Planner developer tools ({PDT}): Reproducible experiments and statistical analysis for developing and testing motion planners,'' in \emph{Proceedings of the Workshop on Evaluating Motion Planning Performance (EMPP), IEEE/RSJ International Conference on Intelligent Robots and Systems (IROS)}, 2022.

\bibitem{gorner2019moveit}
M.~Görner, R.~Haschke, H.~Ritter, and J.~Zhang, ``Moveit! task constructor for task-level motion planning,'' in \emph{IEEE International Conference on Robotics and Automation (ICRA)}, 2019, pp. 190--196.

\bibitem{Diankov2008OpenRAVEAP}
R.~Diankov and J.~J. Kuffner, ``Open{RAVE}: A planning architecture for autonomous robotics,'' in \emph{Carnegie Mellon University}, 2008.

\bibitem{sucan2012open}
I.~A. Sucan, M.~Moll, and L.~E. Kavraki, ``The open motion planning library,'' \emph{IEEE Robotics \& Automation Magazine}, vol.~19, no.~4, pp. 72--82, 2012.

\bibitem{Wilson2024fcit}
T.~S. Wilson, W.~Thomason, Z.~Kingston, L.~E. Kavraki, and J.~D. Gammell, ``Nearest-neighbourless asymptotically optimal motion planning with {Fully} {Connected} {Informed} {Trees} ({FCIT*}),'' in \emph{Proceedings of the {IEEE} International Conference on Robotics and Automation ({ICRA})}, Atlanta, GA, USA, 19--23 May 2025.

\end{thebibliography}

\end{document}